\pdfoutput=1

\documentclass[11pt]{article}

\usepackage[]{acl}

\usepackage{times}
\usepackage{latexsym}

\usepackage{header}

\usepackage[T1]{fontenc}

\usepackage[utf8]{inputenc}

\usepackage{microtype}

\title{Instructions for *ACL Proceedings}

\author{
Linlu Qiu$^1$\thanks{\enskip Equal contribution.}~~\thanks{\enskip Work done as part of the Google AI Residency program.} \quad 
Peter Shaw$^1$\footnotemark[1] \quad 
Panupong Pasupat$^1$ \quad 
Pawe\l{} Krzysztof Nowak$^1$ \\ 
{\bf Tal Linzen}$^{1,2}$ \quad
{\bf Fei Sha}$^1$ \quad
{\bf Kristina Toutanova}$^1$ \\[.5em]

$^1$ Google Research ~~~~~~~~ $^2$ New York Univeristy \\[.5em]
\small{\texttt{\{linluqiu,petershaw,ppasupat,pawelnow,linzen,fsha,kristout\}@google.com}}
}

\title{
Improving Compositional Generalization with Latent Structure \\and Data Augmentation}

\begin{document}

\maketitle
\begin{abstract}

Generic unstructured neural networks
have been shown to struggle on out-of-distribution compositional generalization.
Compositional data augmentation via example recombination has transferred some prior knowledge about compositionality to such black-box neural models for several semantic parsing tasks,
but this often required task-specific engineering or provided limited gains.

We present a more powerful data recombination method using a model called Compositional Structure Learner (CSL). CSL is
a generative model with a quasi-synchronous context-free grammar backbone, which we induce from the training data.
We sample recombined examples from CSL and add them to the fine-tuning data of a pre-trained sequence-to-sequence model (T5). This procedure effectively transfers most of CSL's compositional bias to T5 for diagnostic tasks, and results in a model even stronger than a T5-CSL ensemble on two real world compositional generalization tasks.
This results in new state-of-the-art performance for these challenging semantic parsing tasks requiring generalization to both natural language variation and novel compositions of elements.

\end{abstract}

\section{Introduction}
\label{sec:introduction}

Compositional generalization refers to the ability to generalize to novel combinations of previously observed \emph{atoms}.\footnote{Also commonly referred to as \emph{elements} or \emph{concepts}.} For example, we may ask a model to interpret the instruction ``jump twice'', when the atoms ``jump'' and ``twice'' were each observed separately during training but never in combination with each other~\cite{lake2018generalization}. 

\newcommand{\lispress}[1]{{\fontfamily{cmtt}\selectfont #1}}
\newcommand{\ddone}[1]{{\textcolor{dgreen}{#1}}}
\newcommand{\ddtwo}[1]{{\textcolor{dblue}{#1}}}
\newcommand{\ddthree}[1]{{\textcolor{dred}{#1}}}
\newcommand{\ddfour}[1]{{\textcolor{dyellow}{#1}}}

\renewcommand{\ddone}[1]{\colorbox{orange!10!white}{#1}}
\renewcommand{\ddtwo}[1]{\colorbox{red!8!white}{#1}}
\renewcommand{\ddthree}[1]{\colorbox{blue!8!white}{#1}}
\renewcommand{\ddfour}[1]{\colorbox{dgreen!10!white}{#1}}

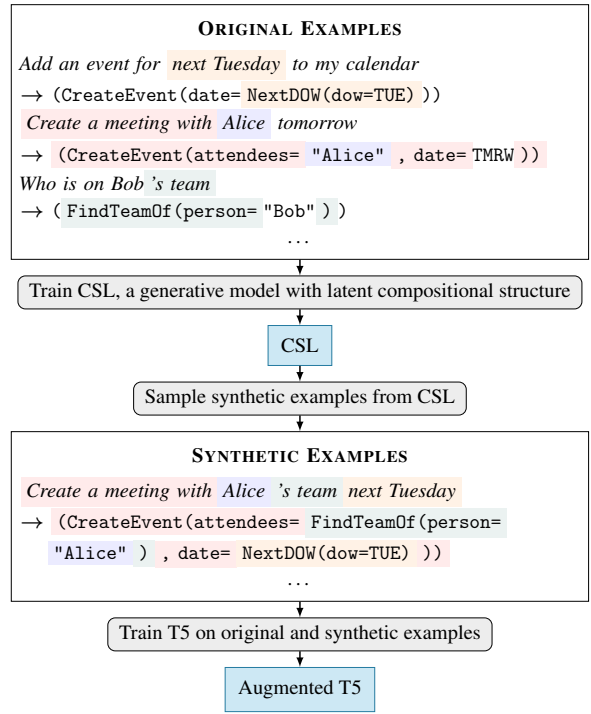
\begin{figure}[!t]
\begin{center}

\scalebox{0.7}{

\begin{tikzpicture} %
\begin{scope}[every node/.style={inner xsep=5pt, inner ysep=5pt}]

\tikzstyle{examples} = [rectangle, draw=black, fill=none, inner ysep=1.8ex]
\tikzstyle{process} = [rectangle, rounded corners, text centered, draw=black, fill=cellshade,inner xsep=6pt, inner ysep=4pt]
\tikzstyle{model} = [rectangle, text centered, draw=dblue, fill=lblue,inner xsep=7pt, inner ysep=7pt]

\tikzstyle{arrow} = [thick,->,>=latex]

\node (n1) [examples, text width=10.5cm] {
{\centering{\textbf{\textsc{Original Examples}}}\\[1ex]}
\textit{Add an event for {\ddone{next Tuesday}} to my calendar} \\
$\boldsymbol{\to}$ \lispress{(CreateEvent(date={\ddone{NextDOW(dow=TUE)}}))} \\

\textit{{\ddtwo{Create a meeting with}}{\ddthree{Alice\vphantom{g}}} tomorrow} \\
$\boldsymbol{\to}$ \lispress{{\ddtwo{(CreateEvent(attendees=}}{\ddthree{"Alice"}}{\ddtwo{, date=}}TMRW{\ddtwo{))}}} \\

\textit{Who is on Bob{\ddfour{'s team}}} \\
$\boldsymbol{\to}$ \lispress{{(\ddfour{FindTeamOf(person=}}"Bob"{\ddfour{)}})}\\

{\centering{\dots}\\}
};

\node (n2) [process, below=0.3cm of n1, align=center] {Train CSL, a generative model with latent compositional structure};

\node (n3) [model, below=0.3cm of n2, align=center] {CSL};

\node (n4) [process, below=0.3cm of n3, align=center] {Sample synthetic examples from CSL};

\node (n5) [examples, text width=10.5cm, below=0.3cm of n4]{
{\centering{\textbf{\textsc{Synthetic Examples}}}\\[1ex]}
\textit{{\ddtwo{Create a meeting with}}{\ddthree{Alice\vphantom{g}}}{\ddfour{'s team\vphantom{g}}}{\ddone{next Tuesday}}} \\
$\boldsymbol{\to}$ \lispress{{\ddtwo{(CreateEvent(attendees=}}{\ddfour{FindTeamOf(person=}} \\
~~~~{\ddthree{"Alice"}}{\ddfour{)}}{\ddtwo{, date=}}{\ddone{NextDOW(dow=TUE)}}{\ddtwo{))}}}\\
{\centering{\dots}\\}
};

\node (n6) [process, below=0.3cm of n5, align=center] {Train T5 on original and synthetic examples};

\node (n7) [model, below=0.3cm of n6, align=center] {Augmented T5};

\draw [arrow] (n1) -- (n2);
\draw [arrow] (n2) -- (n3);
\draw [arrow] (n3) -- (n4);
\draw [arrow] (n4) -- (n5);
\draw [arrow] (n5) -- (n6);
\draw [arrow] (n6) -- (n7);

\end{scope}
\end{tikzpicture}

}

\end{center}

\caption{
An overview of our method for compositional data augmentation with CSL, a generative model with a QCFG backbone, which is automatically induced from the training data. We show a notional set of original and synthetic examples mapping utterances to programs.
}
\label{fig:method_diagram}
\end{figure}

Improving compositional generalization is seen as important for approaching human-like language understanding~\cite{lake2017building, battaglia2018relational} and is practically significant for real world applications, where models deployed in the wild often need to interpret new combinations of elements not well-covered by expensive and potentially skewed annotated training data \cite{herzig2019don, yin2021compositional}.

Generic neural sequence-to-sequence models have improved substantially and reached high levels of performance, particularly when combined with large-scale unsupervised pretraining and sizable in-distribution labeled data. However, these models often perform poorly on out-of-distribution compositional generalization tasks ~\cite{lake2018generalization, furrer2020compositional, shaw-etal-2021-compositional}.  

In contrast,
specialized architectures with discrete latent structure~\cite{chen2020compositional,liu2020compositional,nye2020learning,herzig2020span,shaw-etal-2021-compositional} have made strides in compositional generalization, but without task-specific engineering or ensembling, the gains have been limited to synthetic semantic parsing tasks. Although following SCAN~\cite{lake2018generalization} some increasingly realistic synthetic tasks such as CFQ~\cite{keysers2019measuring} and COGS~\cite{kim-linzen-2020-cogs} have been created, and several approaches achieve good performance on these tasks, the out-of-distribution generalization ability of state-of-the-art models on real-world, non-synthetic tasks is still far from sufficient~\cite{shaw-etal-2021-compositional,yin2021compositional}.

Given their different strengths and weaknesses, it is compelling to combine the compositional bias of such specialized models with the greater flexibility and ability to handle natural language variation that characterizes generic pre-trained neural sequence-to-sequence models. One method for this is \emph{data augmentation}.
For example, \citet{jia2016data} 
generate new training examples
using example recombination via induced high-precision synchronous grammars, resulting in improvements on in-distribution and compositional splits of semantic parsing tasks. Another example is GECA~\cite{andreas-2020-good}, a more general data augmentation approach that does not require task-specific assumptions. GECA achieved further gains on a larger variety of tasks, but provided limited improvements on some compositional generalization challenges.

We present a compositional data augmentation approach that generalizes these earlier methods. Training examples are recombined using the \emph{Compositional Structure Learner} (CSL) model, 
a generative model
with a (quasi-)synchronous context-free grammar (QCFG) backbone, automatically induced from the training data. As illustrated in Figure~\ref{fig:method_diagram}, CSL is used to
sample synthetic training examples, and the union of original and synthesized examples is used to fine-tune the T5 sequence-to-sequence model~\cite{raffel2019exploring}.
CSL is more generally applicable than the method of~\newcite{jia2016data},
employing a generic grammar search algorithm to explore a larger, higher-coverage space of possible grammars. Unlike GECA, CSL can re-combine examples recursively and also defines a probabilistic sampling distribution over input-output pairs.

CSL builds on the NQG model of \citet{shaw-etal-2021-compositional}, a discriminative parsing model over an induced QCFG backbone, which \citet{shaw-etal-2021-compositional} proposed to ensemble with T5. Like NQG, CSL can, on its own, address a variety of compositional generalization diagnostic tasks on synthetic datasets and achieves high precision (but limited recall) on non-synthetic compositional generalization tasks, leading to overall gains when ensembled with T5. However, CSL offers several significant improvements over NQG,
allowing it to efficiently address 
a wider range of datasets (see \S\ref{sec:csl}). Additionally, unlike NQG which is a \emph{discriminative} model assigning probabilities to outputs $y$ given inputs $x$, CSL is a \emph{generative} model which admits sampling from a joint probability distribution $p(x,y)$. This enables the creation of new input-output training examples.

Empirically, augmenting the training data for T5 with samples from CSL transfers most of CSL's compositional bias to T5 for diagnostic tasks (SCAN and COGS), and outperforms a T5+CSL ensemble on non-synthetic compositional generalization tasks defined by compositional splits of GeoQuery~\cite{zelle1996learning} and SMCalFlow~\cite{andreas-etal-2020-task,yin2021compositional}, resulting in new state-of-the-art performance on these splits.\footnote{Our code is available at \url{https://github.com/google-research/language/tree/master/language/compgen/csl}.}
\section{Background and Motivation}
\label{sec:background}

In this section, we discuss the problem setting common to the compositional generalization evaluations we study, and propose some general assumptions that motivate our proposed method.

\paragraph{Problem Setting} Consider a training dataset $\mathcal{D}$ consisting of
input-output pairs $\langle x, y \rangle \in \mathcal{X} \times \mathcal{Y}$, where $\mathcal{X}$ is the set of valid inputs and $\mathcal{Y}$ is the set of valid outputs.
We assume that $\langle x, y \rangle  \in \mathcal{D}$ are sampled from a \emph{source} distribution $p_s(x,y)$. Our model will be evaluated on inputs from a \emph{target} distribution $p_t(x,y)$. 
We make an assumption that the conditional distribution of $y$ given $x$ is unchanged between source and target distributions; i.e., $p_s(y|x) = p_t(y|x)$, which is also a standard assumption for domain adaptation evaluations under covariate shift. Any or all of the following may be true: $p_s(x,y) \neq p_t(x,y)$, $p_s(x) \neq p_t(x)$, $p_s(y) \neq p_t(y)$, and $p_s(x|y) \neq p_t(x|y)$.

What differentiates our setting from other forms of distribution shift is the added assumption that the source and target distributions share common ``atoms'' (see \S\ref{sec:introduction}).
In order to translate this intuitive notion of atom sharing into formal conditions,
we define a general class of models termed \emph{derivational generative models}, based on representing atoms as \emph{functions} which can be recombined via function application. As a modeling hypothesis, we will assume that the training and evaluation distributions can be modeled by derivational generative models that share a common set of functions, but may vary in how they assign probability to derivations formed by recombining these functions.

\begin{figure}[!t]
\begin{center}

\scalebox{0.7}{

\begin{tikzpicture} %
\begin{scope}[every node/.style={inner xsep=5pt, inner ysep=5pt}]

\tikzstyle{func} = [rectangle, rounded corners, text centered, draw=black, fill=none]

\tikzstyle{arrow} = [thick,->,>=latex]

\node (n1) [func, align=center] {
$\lambda \langle x_1, y_1 \rangle . \lambda \langle x_2, y_2 \rangle . \langle x_1$ and $x_2 , y_1~y_2 \rangle $
};

\node (n2) [func, below left=0.3cm and -3.5cm of n1, align=center] {
$\lambda \langle x_1, y_1 \rangle . \langle x_1$ twice $, y_1~y_1 \rangle $
};

\node (n3) [func, below=0.3cm of n2, align=center] {
$\langle$ jump $,$ JUMP $\rangle $
};

\node (n4) [func, below right=0.3cm and -2.5cm of n1, align=center] {
$\langle$ walk $,$ WALK $\rangle $
};

\draw [arrow] (n1) -- (n2);
\draw [arrow] (n2) -- (n3);
\draw [arrow] (n1) -- (n4);

\end{scope}
\end{tikzpicture}

}

\end{center}

\caption{
An example derivation that derives the string pair $\langle$\textit{jump twice and walk} , \textit{JUMP JUMP WALK}$\rangle$.
}
\label{fig:derivation_func}
\end{figure}
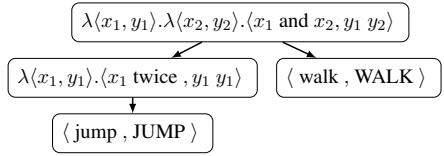

\paragraph{Derivational Generative Models} 
A derivational generative model defines a distribution $p(x,y)$ over input-output pairs.
The model contains a set of functions, $\mathcal{G}$, and a distribution over \emph{derivations}.
A derivation $z$ can be viewed as a tree of functions from $\mathcal{G}$ which \emph{derives} some element $\llbracket z \rrbracket = \langle x,y \rangle \in \mathcal{X} \times \mathcal{Y}$ determined by recursively applying the functions 
in $z$.\footnote{Formally, let $\mathcal{G}$ be a set of partial functions over some set of elements.
If $f \in \mathcal{G}$ is a constant function, then $f$ is a ground term. If $f \in \mathcal{G}$ has arity $k$, and $a_1, \cdots, a_k$ are ground terms, then $f(a_1, \cdots, a_k)$ is a ground term. We define a \emph{derivation} as a ground term that generates an element $\in \mathcal{X} \times \mathcal{Y}$.} An example derivation is shown in Figure~\ref{fig:derivation_func}. %

Given $\mathcal{X}$, $\mathcal{Y}$, and $\mathcal{G}$, we can generate a set $\mathcal{Z}^\mathcal{G}$ of possible derivations.
We define some shorthands for important subsets of $\mathcal{Z}^\mathcal{G}$ for given $x$ and $y$:
\begin{align*}
\mathcal{Z}^\mathcal{G}_{\langle x,y \rangle} &= \{z \in Z^\mathcal{G} \mid  \llbracket z \rrbracket = \langle x,y \rangle \} \\
\mathcal{Z}^\mathcal{G}_{\langle x,* \rangle} &= \{z \in Z^\mathcal{G} \mid  \exists y'\in\mathcal{Y}, \llbracket z \rrbracket = \langle x,y' \rangle \}
\end{align*}

A derivational generative model also consists of some probability distribution $p_{\theta}(z)$ over the set of derivations $\mathcal{Z}^\mathcal{G}$, which we assume to be parameterized by $\theta$. We define $p_{\mathcal{G},\theta}(x,y)$ in terms of  $p_{\theta}(z)$ as:
\begin{equation}\label{eq:joint}
p_{\mathcal{G},\theta}(x,y) = \sum \nolimits_{z \in \mathcal{Z}^\mathcal{G}_{\langle x,y \rangle}} p_{\theta}(z),
\end{equation}
and therefore:
\begin{equation}\label{eq:conditional}
p_{\mathcal{G},\theta}(y|x) = \frac{\sum \nolimits_{z \in \mathcal{Z}^\mathcal{G}_{\langle x,y \rangle}} p_{\theta}(z)}{\sum \nolimits_{z \in \mathcal{Z}^\mathcal{G}_{\langle x,* \rangle}} p_{\theta}(z)},
\end{equation}
for $p_{\mathcal{G},\theta}(x)>0$.

\paragraph{Discussion}
In general, we are interested in a set of functions that captures some knowledge of how the parts of inputs correspond to parts of outputs.
If we can recover some approximation of the underlying set of functions, $\mathcal{G}$, given $\mathcal{D}$, then we could sample derivations consisting of new combinations of functions that are not observed in $\mathcal{D}$. This could potentially help us improve performance on the target distribution, since we assume that the set of functions is unchanged between the source and target distributions, and that what is varying is the distribution over derivations.

However, 
even assuming $\mathcal{G}$ can be exactly recovered given $\mathcal{D}$ is not sufficient to ensure that we can
correctly predict the most likely $y$ given $x$ according to the true $p(y|x)$ (shared between source and target distributions) for $x \sim p_t(x)$.\footnote{One special case is where $|\mathcal{Z}^\mathcal{G}_{\langle x,* \rangle}| = 1$ for all $x$. In this case, every $x$ has exactly one unique derivation and $p_{\mathcal{G}}(y|x)$ is deterministic given $\mathcal{G}$ and does not depend on $\theta$, and therefore recovering $\mathcal{G}$ \emph{is} sufficient.} We must also assume that there exists a parameterization of $p_\theta(z)$ such that when we estimate $\hat{\theta}$ given $\mathcal{D}$, $p_{\mathcal{G},\hat{\theta}}(y|x)$ sufficiently approximates the true $p(y|x)$ for $x \sim p_t(x)$. We hypothesize that conditional independence assumptions with respect to how $p_\theta(z)$ decomposes across the function applications in $z$ can be helpful for this purpose. In particular, such assumptions can enable ``reusing'' conditional probability factors across the exponential space of derivations, potentially improving transfer to the target distribution.

With this intuition in mind, in $\S\ref{sec:method}$ we propose a specific class of functions for $\mathcal{G}$ based on \mbox{(quasi-)synchronous} context-free grammars, as well as a factorization of $p_{\theta}(z)$ with strong conditional independence assumptions.

\section{Proposed Method}
\label{sec:method}

As shown in Figure~\ref{fig:method_diagram}, our method consists of two stages.
First, we induce our generative model, CSL, from training data (\S\ref{sec:csl}). Second, we sample synthetic examples from the generative model and use them to augment the training data for a sequence-to-sequence model (\S\ref{sec:augment}).

\subsection{Compositional Structure Learner (CSL)}
\label{sec:csl}

CSL can be viewed as a derivational generative model, as defined in \S\ref{sec:background}, where the set $\mathcal{G}$ of recursive functions is defined by a (quasi-)synchronous context free grammar (QCFG).\footnote{QCFG rules can be interpreted as functions which are limited to string concatenation. For notational convenience, we will therefore treat $\mathcal{G}$ as a set of QCFG rules in \S\ref{sec:method}.} %

We first describe the grammar formalism and the parameterization of our probabilistic model. Then we describe our two-stage learning procedure for inducing a grammar and learning the model parameters. CSL builds on the NQG model of ~\citet{shaw-etal-2021-compositional}, with several key differences discussed in the following sections:
\begin{itemize}
  \item Unlike NQG, which is discriminative, CSL is a \emph{generative} model that admits efficient sampling from the joint distribution $p(x,y)$.
  \item CSL enables a more expressive set of grammar rules than NQG. %
  \item CSL offers a more computationally efficient and parallelizable grammar induction algorithm and a more efficient parametric model, allowing the method to scale up to larger datasets such as SMCalFlow-CS.
\end{itemize}
See section \ref{sec:analysis} for experiments and analysis comparing the components of CSL and NQG. 

\subsubsection{Grammar Formalism}
\label{sec:grammar-formalism}

\begin{figure}[!t]
\begin{center}

\scalebox{0.7}{

\begin{tikzpicture} %
\begin{scope}[every node/.style={inner xsep=5pt, inner ysep=5pt}]

\tikzstyle{func} = [rectangle, rounded corners, text centered, draw=black, fill=none]

\tikzstyle{arrow} = [thick,->,>=latex]

\node (n1) [func, align=center] {
$\mbox{NT} \rightarrow \langle \mbox{NT}_\boxnum{1} $ and $\mbox{NT}_\boxnum{2} , \mbox{NT}_\boxnum{1}~ \mbox{NT}_\boxnum{2} \rangle$

};

\node (n2) [func, below left=0.3cm and -3.5cm of n1, align=center] {
$\mbox{NT} \rightarrow \langle \mbox{NT}_\boxnum{1}$ twice$, \mbox{NT}_\boxnum{1} \mbox{NT}_\boxnum{1} \rangle$
};

\node (n3) [func, below=0.3cm of n2, align=center] {
$\mbox{NT} \rightarrow \langle$ jump $,$ JUMP $ \rangle$
};

\node (n4) [func, below right=0.3cm and -2.5cm of n1, align=center] {
$\mbox{NT} \rightarrow \langle $walk$,$ WALK$ \rangle$
};

\draw [arrow] (n1) -- (n2);
\draw [arrow] (n2) -- (n3);
\draw [arrow] (n1) -- (n4);

\end{scope}
\end{tikzpicture}

}

\end{center}

\caption{
The example derivation of Figure \ref{fig:derivation_func} using QCFG notation.
}
\label{fig:derivation_qcfg}
\end{figure}
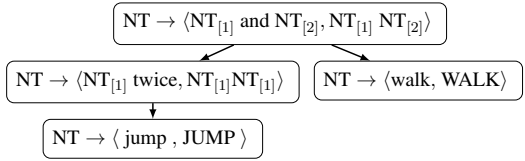

An example QCFG derivation is shown in Figure~\ref{fig:derivation_qcfg}, and the notation for QCFGs is reviewed in Appendix~\ref{sec:appendix-model-qcfg}. Notably, the correspondence between rules over input and output strings in QCFGs is akin to a homomorphism between syntactic and semantic structures, commonly posited by formal theories of compositional semantics~\cite{montague1970universal,janssen1997compositionality}.
We restrict our grammars to have only a single unique nonterminal symbol, $NT$. In constrast to standard synchronous context-free grammars (SCFGs), our grammars can be \emph{quasi-synchronous}~\cite{smith2006quasi} because we allow a one-to-many alignment between non-terminals.\footnote{Concretely, for a rule $NT \rightarrow \langle \alpha,\beta \rangle $, a non-terminal in $\alpha$ can share an index with more than one non-terminal in $\beta$. This is important for datasets such as SCAN, as it allows rules such as $\mbox{NT} \rightarrow \langle \mbox{NT}_\boxnum{1} \text{twice}, \mbox{NT}_\boxnum{1} \mbox{NT}_\boxnum{1} \rangle$ which enable repetition.}
Unlike the formalism of~\citet{shaw-etal-2021-compositional}, which limited rules to contain $\leq 2$ non-terminals, in the current work the maximal number of non-terminals is a configurable parameter,\footnote{We find that $4$ is a computationally tractable choice for the datasets we study.} which enables inducing grammars with higher coverage for certain datasets.

\subsubsection{Probabilistic Model}
\label{sec:method-parameterization}

We factorize the probability of a derivation in terms of conditional probabilities of sequentially expanding a rule from its parent.
Formally, let $r$ denote a rule expanded from its parent rule $r_p$'s $NT_{\boxnum{i}}$ nonterminal (or a special symbol at the root of the derivation tree).\footnote{Using the example derivation from Figure ~\ref{fig:derivation_qcfg}, for the rule application $r = \mbox{NT} \rightarrow \langle \mbox{walk},\mbox{WALK}\rangle$, we have
$r_p = \mbox{NT} \rightarrow \langle \mbox{NT}_\boxnum{1} \mbox{ and } \mbox{NT}_\boxnum{2}, \mbox{NT}_\boxnum{1} \mbox{NT}_\boxnum{2} \rangle$
and the expansion probability for that rule application is $p(r|r_p,2)$.}
We assume conditional independence and factorize the probability of $z$ as:
\begin{equation} \label{eq:derivation_prob}
p_{\theta}(z) = \prod \limits_{r, r_p, i \in z} p_{\theta}(r|r_p,i)
\end{equation}
This non-terminal annotation with context from the tree is akin to parent annotation or other structure conditioning for probabilistic context-free grammars~\cite{johnson98,klein-manning-2003-accurate}.

Using independent parameters for each combination of a rule, its parent, and non-terminal index may lead to overfitting to the training set, limiting our ability to generalize to new combinations of rule applications that are needed for compositional generalization. We therefore factor this distribution using a soft clustering into a set of \emph{latent states} $\mathcal{S}$  representing parent rule application contexts:

\begin{equation}\label{eq:rule}
p_{\theta}(r|r_p,i) =  \sum \limits_{s \in \mathcal{S}} p_{\theta}(r|s)p_{\theta}(s|r_p,i)
\end{equation}
where $p_{\theta}(s|r_p,i) \propto e^{\theta_{r_p,i,s}}$ and 
$p_{\theta}(r|s) \propto e^{\theta_{s,r}}$ and the $\theta$s are scalar parameters.\footnote{The full definition of these terms is in Appendix~\ref{sec:appendix-model-parameterization}.}
The number of context states $|\mathcal{S}|$ is a hyperparameter. While these conditional independence assumptions may still be too strong for some cases (see Appendix \ref{sec:appendix-analysis-states}), we find them to be a useful approximation in practice. We also optionally consider a task-specific output CFG, which defines valid output constructions.\footnote{The outputs for several of the tasks we study consist of executable programs or logical terms, for which we can assume the availability of a CFG for parsing. Details are in Appendix~\ref{sec:appendix-datasets}.}

\subsubsection{Learning Procedure}
A principled method to estimate $\mathcal{G}$ and $\theta$ given $\mathcal{D}$ would be to find the MAP estimate based on some prior, $p(\mathcal{G},\theta)$, that encourages compositionality:
\begin{equation}\label{eq:map}
\underset{\mathcal{G},\mathcal{\theta}}{\argmax}
~~p(\mathcal{G},\theta) \times \!\!\prod \limits_{\langle x,y \rangle \in \mathcal{D}} p_{\mathcal{G},\mathcal{\theta}}(x,y)
\end{equation}

However, since optimizing $\mathcal{G}$ and $\theta$ jointly is computationally challenging, we adopt a two-stage process similar to that of \citet{shaw-etal-2021-compositional}.
First, we learn an unweighted grammar using a surrogate objective for the likelihood of the data and a compression-based compositional prior that encourages smaller grammars that reuse rules in multiple contexts, inspired by the Minimum Description Length (MDL) principle~\cite{rissanen1978modeling, grunwald2004tutorial}. We describe the induction objective and algorithm in \S\ref{sec:induction}.
Second, given an unweighted grammar $\mathcal{G}$, we optimize the parameters $\theta$ by maximizing the log-likelihood of $p_{\mathcal{G},\theta}(x,y)$, as defined by Eq.~\ref{eq:joint}, using the Adam optimizer~\cite{kingma2014adam}.\footnote{To optimize $\theta$ efficiently, we use a variant of the CKY algorithm~\cite{cocke1969programming,kasami1965,younger1967recognition} to determine the set of derivations, represented as a parse forest, and use dynamic programming to efficiently sum over this set.}

\subsubsection{Grammar Induction Algorithm}
\label{sec:induction}

Our method for inducing a QCFG is based on that of \citet{shaw-etal-2021-compositional}, but with several modifications, which improve the computational scalability of the algorithm as well as the precision and coverage of the induced grammar. We analyze the relative performance of the two algorithms in~\S\ref{sec:analysis}.

\paragraph{Objective}

The main idea of the grammar induction objective, $L_\mathcal{D}(\mathcal{G})$, is to balance the size of the grammar with its ability to fit the training data:
\begin{equation}\label{eq:grammar_objective}
L_\mathcal{D}(\mathcal{G}) = \sum\limits_{NT \rightarrow \langle \alpha, \beta \rangle \in \mathcal{G}} |\alpha| + |\beta| - c_\mathcal{D}(\alpha, \beta),
\end{equation}
where $|\!\cdot\!|$ is a weighted count of terminal and nonterminal tokens (the relative cost of a terminal vs. nonterminal token is a hyperparameter)
and
\begin{equation}\label{eq:grammar_objective2}
c_\mathcal{D}(\alpha, \beta) = k_{\alpha}  \ln \hat{p}_\mathcal{D}(\alpha | \beta) + k_{\beta} \ln \hat{p}_\mathcal{D}(\beta | \alpha)
\end{equation}
where $k_\alpha$ and $k_\beta$ are hyperparameters and $\hat{p}_\mathcal{D}(\alpha | \beta)$ is equal to the fraction of examples $\langle x,y \rangle \in \mathcal{D}_{train}$ where $\alpha$ ``occurs in'' $x$ out of the examples where $\beta$ ``occurs in'' $y$,
and vice versa for $\hat{p}_\mathcal{D}(\beta | \alpha)$.\footnote{By $\alpha$ ``occurs in'' $x$, we mean that there exists some substitution for any non-terminals in $\alpha$ such that it is a substring or equal to $x$.}
The correlation between $\alpha$ and $\beta$ as measured by the $\hat{p}$ terms provides a measure related to how well the rule fits the training data.
We use sampling to optimize the computation of $\hat{p}_\mathcal{D}$ for larger datasets. While conceptually similar to the objective used by NQG, we found that CSL's objective is more efficient to compute and can be more effective at penalizing rules that lead to lower precision.\footnote{For example, CSL’s objective enables our algorithm to induce a ``clean'' 20 rule grammar for SCAN, while using our algorithm with the objective of NQG leads to grammars with additional spurious rules for SCAN.}

\paragraph{Initialization} To initialize $\mathcal{G}$ we add a rule $NT \rightarrow \langle x,y \rangle$ for every $\langle x,y \rangle \in \mathcal{D}$. We also optionally add a set of seed rules, such as $NT \rightarrow \langle x,x \rangle$ where a terminal or span $x$ is shared between the input and output vocabularies. For details on seed rules used for each dataset, see Appendix~\ref{sec:appendix-datasets}.

\paragraph{Greedy Algorithm} Following the initialization of the set of rules $\mathcal{G}$, we use an approximate parallel greedy search algorithm to optimize $L_\mathcal{D}(\mathcal{G})$, while maintaining the invariant that all examples in $\mathcal{D}$ can be derived by $\mathcal{G}$. 

At each iteration, the algorithm considers each rule $r$ in the current grammar in parallel. The algorithm determines a (potentially empty) set of candidate actions for each $r$. Each candidate action consists of adding a new rule to the grammar that can be combined with an existing rule to derive $r$, enabling $r$ to be removed. Certain candidate actions may enable removing other rules, too. The algorithm then selects the candidate action that leads to the greatest improvement in the induction objective, if any action exists that leads to an improvement. The selected actions are then aggregated and executed, resulting in a new set of rules. The algorithm continues until no further actions are selected, or a maximum number of steps is reached. The detailed implementation of the greedy algorithm is detailed in Appendix~\ref{sec:appendix-model}.

\subsubsection{Inference Procedure}
\label{sec:inference-procedure}

While the primary goal of CSL is to be used to sample new examples for data augmentation (discussed next), we can also use CSL as a  discriminative parsing model, by using a %
variant of the CKY algorithm to find the highest scoring derivation $z$ that maximizes Eq.~\ref{eq:derivation_prob} for a given input $x$. We then output the corresponding $y$ if it can be derived by the given output CFG, or %
if no output CFG is provided.

\subsection{Data Augmentation}
\label{sec:augment}

We synthesize a configurable number of examples by sampling from the learned generative model, CSL.\footnote{For all experiments, we sample 100,000 synthetic examples unless otherwise indicated.}
To generate a synthetic example $(x, y)$, we use forward sampling:  we start from the single $NT$ symbol and sample recursively to expand each nonterminal symbol with a rule, based on $p_{\theta}(r|r_p,i)$ defined by Eq.~\ref{eq:rule}.\footnote{We ensure the sampled $y$ can be generated by a CFG defining valid outputs, if one is provided for the given task, by sampling from the intersection of $\mathcal{G}$ and the provided output CFG.}

Given that generic sequence-to-sequence models perform poorly on length extrapolation~\citep{newman2020the}, we optionally bias our sampling to favor deeper derivations. We achieve this by adding a bias $\delta > 0$ to $\theta_{t,r}$ for any rule $r$ that contains more nonterminals than our configurable threshold.

We fine-tune T5 on the union of the original training data and the synthesized data. Following ~\citet{jia2016data}, we ensure an approximately equal number of original and synthesized examples are used for training. We achieve this by %
replicating original or synthetic examples as needed.

\section{Experiments and Analysis}
\label{sec:experiments}

In this section, we comparatively evaluate and analyze our main proposed method, T5+CSL-Aug., which uses CSL to generate examples for augmenting the training data of T5.

\begin{table*}[!t]
\begin{center}

\scalebox{0.74}{
\begin{tabular}{lccccccccccccccc}
\toprule
 & \multicolumn{4}{c}{{\bf{\textsc{SCAN}}}}
 & \multicolumn{1}{c}{{\bf{\textsc{COGS}}}} 
 & \multicolumn{4}{c}{{\bf{\textsc{GeoQuery}}}} 
 & \multicolumn{6}{c}{{\bf\textsc{SMCalFlow-CS}}} \\
 \cmidrule(lr){2-5} \cmidrule(lr){6-6} \cmidrule(lr){7-10} \cmidrule(lr){11-16}
System & Jump & Left & Len. & MCD & Gen. & Std. & Templ. & TMCD & Len. & 8-S & 8-C & 16-S & 16-C & 32-S & 32-C \\
\midrule
NQG-T5 & 
\bf 100.0 & \bf 100.0 & \bf 100.0 & \bf 100.0 & 97.9 & \bf 92.9 & 84.2 & 71.4 & 53.9 & --- & --- & --- & --- & --- & --- \\
SpanBasedSP & 
--- & --- & --- & --- & --- & 78.9 & 76.3 & 56.5 & 53.9 & --- & --- & --- & --- & --- & --- \\
LeAR & --- & --- & --- & --- & 97.7 & --- & --- & --- & --- & --- & --- & --- & --- & --- & --- \\
C2F & 
--- & --- & --- & --- & --- & --- & --- & --- & --- & --- & --- & 83.0 & 40.6 & 83.6 & 54.6 \\
C2F+SS & 
--- & --- & --- & --- & --- & --- & --- & --- & --- & --- & --- & \bf 83.8 & 47.4 & 83.7 & 61.9 \\
\midrule
T5 & 
\bf 99.5 & 62.0 & 14.4 & 15.4 & 89.8 & \bf 92.9 & 84.8 & 69.2 & 41.8 & \bf 84.7 & 34.7 & \bf 84.7 & 44.7 & \bf 85.2 & 59.0 \\
T5+GECA & 
\bf 99.7 & 57.6 & 10.5 & 22.8 & --- & 92.5 & 82.8 & 66.5 & 45.8 & --- & --- & --- & --- & --- & --- \\
\midrule
T5+CSL-Aug. & 
\bf 99.7 & \bf 100.0 & \bf 99.2 & \bf 99.4 & \bf 99.5 & \bf 93.3 & \bf 89.3 & \bf 74.9 & \bf 67.8 & 83.5 & \bf 51.6 & 83.4 & \bf 61.4 & 84.0 & \bf 70.4 \\
\bottomrule
\end{tabular}
}
\caption{{\bf Main Results.} We compare the performance of our proposed method, T5+CSL-Aug., to prior work across synthetic (SCAN, COGS) and non-synthetic (GeoQuery, SMCalFlow-CS) tasks. Boldfaced results are within 1.0 points of the best result. 
}
\label{tab:main_joint}
\end{center}
\end{table*}

\subsection{Datasets}
\label{sec:datasets}

We evaluate our approach on both synthetic benchmarks designed for controlled assessments of compositional generalization,
and non-synthetic evaluations, which introduce the additional challenge of handling natural language variation. For example, some words in the test data might never appear during training. %
Further details on datasets and preprocessing are in Appendix~\ref{sec:appendix-datasets}.

\paragraph{SCAN}
The SCAN dataset contains navigation commands paired with action sequences. We consider three compositional data splits from \citet{lake2018generalization}: the \emph{jump} and \emph{turn left} splits (where a new primitive is used in novel combinations), and the \emph{length} split.
We also consider the \emph{MCD} splits from \citet{keysers2019measuring} created by making the distributions of compositional structures in training and test data as divergent as possible.

\paragraph{COGS}
The COGS dataset \cite{kim-linzen-2020-cogs} contains sentences paired with logical forms.
We use the generalization test set, which tests generalization to novel linguistic structures.
As SCFGs cannot handle logical variables \cite{wong-mooney-2007-learning},
we convert the outputs into equivalent variable-free forms.

\paragraph{GeoQuery}
 GeoQuery~\cite{zelle1996learning,tang2001using} contains human-authored questions paired with meaning representations. We report results on the standard data split as well as three compositional splits based on those introduced
in \citet{shaw-etal-2021-compositional}: the \emph{template} split (where abstract output templates in training and test data are disjoint~\cite{finegan2018improving}), the \emph{TMCD} split (an extension of MCD for non-synthetic data), and the \emph{length} split.\footnote{For GeoQuery, to reduce variance due to small dataset sizes, we average all results over 3 runs. For the Template and TMCD splits we additionally average over 3 splits generated with different random seeds. Variance is reported in Appendix~\ref{sec:appendix-analysis-geoquery}}

\paragraph{SMCalFlow-CS}
\citet{yin2021compositional} proposed a \emph{compositional skills} split of  SMCalFlow~\cite{andreas-etal-2020-task} that contains single-turn sentences from one of two domains related to creating calendar events or querying an org chart, paired with LISP programs. The single-domain (S) test set has examples from a single domain, while the cross-domain (C) test set has sentences that require knowledge from both domains (e.g., ``create an event with my manager''). Only a small number of cross-domain examples (8, 16, or 32) are seen during training.

\subsection{Baselines}
\label{sec:baselines}

Our primary goal is to evaluate T5+CSL-Aug.
in comparison to T5 and T5+GECA, a method augmenting training data with GECA which is prior state of the art for data augmentation~\cite{andreas-2020-good}.\footnote{For all experiments with T5, we show results for T5-Base (220M parameters). Prior work found T5-Base to perform best on the compositional splits of SCAN and GeoQuery~\cite{furrer2020compositional,shaw-etal-2021-compositional}. We found a similar trend for COGS and SMCalFlow-CS after evaluating T5-Large and T5-3B.} Details and hyperparameters for the GECA experiments are available in Appendix \ref{sec:appendix-analysis-geca}. We also compare with representative prior state-of-the-art methods.
For SCAN,  NQG-T5~\cite{shaw-etal-2021-compositional} is one of several specialized models that achieves 100\% accuracy across multiple splits~\cite{chen2020compositional,liu2020compositional,nye2020learning,herzig2020span}. For COGS, we show results from LeAR~\cite{liu2021learning}, the previously reported state-of-the-art on COGS.\footnote{We do not show LeAR results for SCAN and GeoQuery as ~\citet{liu2021learning} did not report results for SCAN and reported GeoQuery results using a different template split and a different evaluation metric.} We also report new results for NQG-T5 on COGS.
For GeoQuery, we report results for NQG-T5\footnote{Some of our results for NQG-T5 are different than those reported in \citet{shaw-etal-2021-compositional} as we average over 3 new GeoQuery template and TMCD splits, as described in \S\ref{sec:datasets}.} and SpanBasedSP~\cite{herzig2020span} on the GeoQuery splits we study.\footnote{We report exact match accuracy for SpanBasedSP, as opposed to denotation accuracy reported in ~\citet{herzig2020span}.} For SMCalFlow-CS, we show the strongest previously reported results by \citet{yin2021compositional}, which include
the coarse2fine (C2F) model of  \citet{dong2018coarse} as a baseline, as well C2F combined with the span-supervised (SS) attention method of \citet{yin2021compositional}. We found it was not computationally feasible to run NQG-T5 on SMCalFlow.

\begin{table*}[!t]
\begin{center}

\scalebox{0.74}{
\begin{tabular}{lccccccccccc}
\toprule
 & 
 & \multicolumn{3}{c}{{$x \in \mathcal{X}_{CSL}$}} 
 & \multicolumn{3}{c}{{$x \notin \mathcal{X}_{CSL}$}} 
 & \multicolumn{4}{c}{{All}} \\
 
 \cmidrule(lr){3-5} \cmidrule(lr){6-8} \cmidrule(lr){9-12}
 
Dataset & $\%^{\mathcal{X}_{CSL}}$ & T5 & CSL  & Aug. & T5 & CSL  & Aug. & T5 & CSL & Ens. & Aug. \\
\midrule
GeoQuery Templ. & 61.0 & 93.1 & 96.6 & \bf 97.1 & 71.6 & 0.0 & \bf 76.9 & 84.8 & 58.9 & 86.9 & \bf 89.3 \\
GeoQuery TMCD   & 44.3 & 88.4 & 90.3 & \bf 93.9 & 53.8 & 0.0 & \bf 59.9 & 69.2 & 39.9 & 70.0 & \bf 74.9 \\
GeoQuery Length & 29.0 & 51.2 & \bf 91.6 & 83.6 & 35.4 & 0.0 & \bf 61.3 & 40.0 & 26.6 & 51.7 & \bf 67.8 \\
\midrule
SMCalFlow-CS 8-C  & 6.6  & 52.3 & \bf 79.6 & 72.7 & 33.4 & 0.0 & \bf 50.1 & 34.7 & 5.3 & 36.5 & \bf 51.6 \\
SMCalFlow-CS 16-C & 11.6 & 59.7 & \bf 84.4 & \bf 84.4 & 42.7 & 0.0 & \bf 58.4 & 44.7 & 9.8  & 47.5 & \bf 61.4 \\
SMCalFlow-CS 32-C & 13.0 & 74.4 & 87.2 & \bf 88.4 & 56.7 & 0.0 & \bf 67.8 & 59.0 & 11.3 & 60.6 & \bf 70.4 \\
\bottomrule
\end{tabular}
}
\caption{We compare T5, CSL used as a parsing model, and T5+CSL-Aug. (abbreviated as Aug.). 
We partition evaluation examples by whether CSL generates an output ($x \in \mathcal{X}_{CSL}$) or not ($x \notin \mathcal{X}_{CSL}$) for a given input. The percentage of examples in the former subset ($\%^{\mathcal{X}_{CSL}}$) is limited by the coverage of the induced grammar (see Section~\ref{sec:inference-procedure}). We also compare with an ensemble of T5 and CSL (Ens.) similar to that of \citet{shaw-etal-2021-compositional}, where we use the output of CSL if $x \in \mathcal{X}_{CSL}$ and otherwise fall back to T5.}
\label{tab:quadrant_analysis_small}
\end{center}
\end{table*}

\subsection{Main Results}
\label{sec:experiments-results}

The results are shown in 
Table~\ref{tab:main_joint}.
For synthetic datasets, the induced grammars have high coverage, making the CSL model highly effective for data augmentation.
When we use CSL to generate additional training data for T5 (T5+CSL-Aug.), 
the performance of T5 improves to nearly solving SCAN and achieving state-of-the-art on COGS.

For non-synthetic tasks, %
T5+CSL-Aug.\@ leads to new state-of-the-art accuracy on all compositional splits. However, performance is slightly worse on the single-domain splits of SMCalFlow-CS. Based on error analysis in Appendix~\ref{sec:appendix-analysis-smcalflow}, we find a significant degree of inherent ambiguity for the remaining errors on the single-domain split, which may contribute to this result.

Using CSL for data augmentation outperforms using GECA on SCAN and GeoQuery. We did not find it computationally feasible to run GECA on COGS or SMCalFlow-CS. On some splits, using GECA to augment the training data for T5 can lead to worse performance, as GECA can over-generate incorrect examples. We provide further analysis comparing CSL and GECA in Appendix~\ref{sec:appendix-analysis-geca}.

\subsection{Analysis and Discussion}
\label{sec:analysis}

The performance of T5+CSL-Aug.\@ is dependent on the CSL grammar backbone, parametric model, and data sampling details.  We analyze the accuracy of T5+CSL-Aug.\@ in relation to CSL's coverage, and summarize the impact of the design choices in CSL that depart from prior work.

\paragraph{Performance Breakdown}

CSL provides analyses for and can only sample inputs  covered by its grammar $x \in \mathcal{X}_{CSL}$, which is often a strict subset of all possible utterances. It is therefore interesting to see how data augmentation impacts T5’s performance on covered and non-covered inputs.

In Table~\ref{tab:quadrant_analysis_small}, we analyze the relative performance of T5, CSL, and combinations of T5 and CSL using ensembling and data augmentation, for non-synthetic compositional splits,  partitioning inputs based on whether they are covered by CSL (the same analysis for all splits can be found in Appendix~\ref{sec:appendix-analysis-breakdown}).
An ensemble model can help T5 only when $x \in \mathcal{X}_{CSL}$, but we can see from Table~\ref{tab:quadrant_analysis_small} that data augmentation improves model performance even on inputs not covered by the grammar.  For example, for the GeoQuery Length split, performance on non-covered inputs improves from 35.3 to 60.9. This means that T5 is  generalizing from the sampled data ($x \in \mathcal{X}_{CSL}$) to $x \notin \mathcal{X}_{CSL}$.

\eat{
In \fs{Maybe streamline some abbreviation here: (1) always T5+CSL-Aug refers to csl+augmentation, (2) use T5\& csl refers to ensembling without data augmentation? In the caption of Table 2, "T5+CSL using data augmentation" is used to refer to T5+CSL-aug, which has been used.}Table~\ref{tab:quadrant_analysis_small}, we analyze the relative performance of T5, CSL, and combinations of T5 and CSL using ensembling and data augmentation, for non-synthetic compositional splits,\footnote{The results for all splits can be found in Appendix~\ref{sec:appendix-analysis-breakdown}.} partitioning inputs based on whether they are covered by CSL.
CSL can produce outputs only when the input is covered by the grammar, while T5 trained on original training data performs well on in-distribution inputs but struggles for out-of-distribution ones. An ensemble model works well on the inputs covered by either CSL or T5, but cannot generalize beyond such inputs.

Our data augmentation procedure is akin to \emph{knowledge distillation}, distilling the ability to compositionally generalize to out-of-distribution inputs from the teacher model CSL to the student model T5.
With this data augmentation, T5+CSL-Aug. makes accuracy gains on inputs both within and outside the scope of the grammar.
This means that the model is, to some degree, generalizing from the distillation data ($x \in \mathcal{X}_{CSL}$) to $x \notin \mathcal{X}_{CSL}$.
\kt{It would be nice to add a sentence with one  example of this generalization phenomenon, citing concrete numbers from the Table.}
}

\paragraph{Comparison with NQG} We cannot compare using CSL for data augmentation directly with using its closely related predecessor NQG~\cite{shaw-etal-2021-compositional} for data augmentation, as NQG is a discriminative parsing model and not a probabilistic generative model that enables sampling new examples. However, we include comparisons of the novel components of CSL relative to the related components of NQG in the following sections, which analyze CSL’s grammar induction algorithm and parametric model. 

\paragraph{Grammar Induction}
The grammar induction algorithm of CSL is significantly more scalable than that of NQG, enabling more than 90\% decrease in runtime for GeoQuery, and enabling induction to scale to larger datasets such as SMCalFlow-CS. CSL can also induce higher coverage grammars than NQG in some cases, while maintaining high precision. For example, for COGS, the grammar induced by CSL can derive 99.9\% of the evaluation set while the grammar induced by NQG can only derive 64.9\%. Appendix~\ref{sec:appendix-analysis-nqg} contains further analysis comparing the grammar induction algorithms of CSL and NQG. Of course, the grammars induced by CSL can still lack coverage for some datasets, as shown in Table~\ref{tab:quadrant_analysis_small}. We analyze the limitations of QCFGs in Appendix~\ref{sec:qcfg-limitations}.

\paragraph{Parameteric Model}
We find that the simple parameteric model of CSL performs comparably in terms of parsing accuracy to the BERT-based discriminative model of NQG given the same grammar (see Appendix~\ref{sec:appendix-analysis-nqg}). It is also more scalable because it does not require the computation of a partition function. The variable number of state clusters (\S\ref{sec:method-parameterization}) provides a powerful knob for tuning the amount of context sensitivity (see Appendix~\ref{sec:appendix-analysis-states}) to sufficiently fit the training data while also extrapolating to out-of-distribution compositions. We believe further improvements to the parametric model (e.g. using pre-trained representations) have strong potential to improve overall accuracy.

\paragraph{Sampling} Results on most splits are significantly improved by using CSL's parametric model compared to sampling uniformly from the induced grammar (Appendix~\ref{sec:appendix-analysis-temp}), pointing to a potential source of gains over unweighted augmentation approaches like GECA. However, for SMCalFlow-CS, a higher sampling temperature can improve performance, especially on the 8-shot split, as it leads to $>15$ times the number of cross-domain examples being sampled, given their low percentage in the training data. Determining improved methods for biasing the sampling towards examples most relevant to improving performance on the target distribution is an important direction. In Appendix~\ref{sec:appendix-analysis-unlabeled} we explore a setting where we assume access to unlabeled examples from the target distribution, and use these to update the parametric model. We find that this improves sample efficiency with respect to the number of sampled synthetic examples, but can have minimal effect when a sufficiently large number of examples can be sampled. We believe this is a promising research direction.

\section{Related Work}

\paragraph{Grammar Induction}
Before the trend towards sequence-to-sequence models, significant prior work in semantic parsing explored inducing SCFG~\cite{wong-mooney-2006-learning,wong-mooney-2007-learning,andreas-etal-2013-semantic} and CCG~\cite{zettlemoyer2005learning,zettlemoyer2007online,kwiatkowski2010inducing,kwiatkowski2013scaling,artzi-etal-2014-learning} grammars of the input-output pairs. SCFGs have also been applied to machine translation \cite{chiang2007hierarchical,blunsom-etal-2008-discriminative,saers-etal-2013-unsupervised}. Compression-based objectives similar to ours have also been applied to CFG induction \cite{grunwald1995minimum}. Recently, the method of \citet{kim2021sequence} learns neural parameterized QCFG grammars, which can avoid the pitfalls in coverage of lexicalized grammars such as the ones we learn; however the approach can be computationally demanding for longer input-output pairs. 

\paragraph{Data Augmentation} Data augmentation has been widely used for semantic parsing and related tasks~\cite{jia2016data, andreas-2020-good, Akyrek2021LearningTR, Wang2021LearningTS, Zhong2020GroundedAF, Oren2021FindingNI, Tran2020GeneratingSD, Guo2020SequencelevelMS, Guo2021RevisitingIB}. \citet{jia2016data} perform data recombination using an induced SCFG but their approach requires domain-specific heuristics.  GECA~\cite{andreas-2020-good} provides a more general solution, which we analyzed in \S\ref{sec:analysis}. 
The method of \citet{Akyrek2021LearningTR} is appealing because it can learn data recombinations without committing to a grammar formalism, although gains were limited relative to symbolic methods. The recombination approach of \citet{Guo2020SequencelevelMS} demonstrates gains for translation tasks but is not as effective as GECA for semantic parsing tasks.
Other approaches leverage a forward semantic parser and a backward input generator with some variants~\cite{Wang2021LearningTS, Zhong2020GroundedAF, Tran2020GeneratingSD, Guo2021RevisitingIB}, but most of these approaches do not explicitly explore the compositional generalization setting. 
\citet{Oren2021FindingNI} propose an approach to sample more structurally-diverse data to improve compositional generalization, given a manually specified SCFG.

\paragraph{Compositional Generalization}
Beyond data augmentation, many approaches have been pursued to improve compositional generalization in semantic parsing, including model architectures~\cite{li2019compositional, russin2019compositional, gordon2020permutation, liu2020compositional,nye2020learning, chen2020compositional, zheng2020compositional, oren-etal-2020-improving,herzig2020span, ruiz2021iterative,wang2021structured}, different Transformer variations~\cite{csordas2021devil, ontanon2021making}, ensemble models~\cite{shaw-etal-2021-compositional}, intermediate representations~\cite{herzig2021unlocking}, meta-learning~\cite{lake2019compositional, conklin2021meta, zhu2021learning}, and auxiliary objectives to bias attention in encoder-decoder models~\cite{yin2021compositional,jiang2021inducing}. Also, \citet{furrer2020compositional} compared pre-trained models with specialized architectures.

\section{Conclusion}

We showed that the Compositional Structure Learner (CSL) generative model improves the state of the art on compositional generalization challenges for two real-world semantic parsing datasets when used to augment the task training data for the generic pre-trained T5 model. Data augmentation using CSL was also largely sufficient to distill CSL's knowledge about compositional structures into T5 for multiple synthetic compositional generalization evaluations. While CSL has limitations (notably, the QCFG formalism is not a good fit for all phenomena in the mapping of natural language to corresponding logical forms), our experiments suggest the strong potential of more powerful probabilistic models over automatically induced latent structures as data generators for black-box pretrained sequence-to-sequence models.

\section*{Acknowledgements}

We thank Nitish Gupta, Ming-Wei Chang, William Cohen, Pengcheng Yin, Waleed Ammar, the Google Research Language team, and the anonymous reviewers for helpful comments and discussions.
\section*{Ethical Considerations}

This paper proposed methods to improve compositional generalization in semantic parsing. While we hope that improvements in compositional generalization would lead to systems that generalize better to languages not well represented in small training sets, in this work we have only evaluated our methods on semantic parsing datasets in English.

\bibliography{anthology,custom}
\bibliographystyle{acl_natbib}

\clearpage

\appendix\section*{Appendix}

The appendix is organized into three sections:
\begin{itemize}
   \item Appendix~\ref{sec:appendix-datasets} contains dataset and preprocessing details.
   \item Appendix~\ref{sec:appendix-model} contains modeling details and hyperparameters.
   \item Appendix~\ref{sec:appendix-analysis} contains additional experiments and analysis.
\end{itemize}

\section{Dataset and Preprocessing Details}
\label{sec:appendix-datasets}

\begin{table}[h!]
\centering
\resizebox{0.95\columnwidth}{!}{
\begin{tabular}{@{}llccc@{}}
\toprule
Dataset                       & Split     & Train & Dev & Test \\
\midrule      
\multirow{6}{*}{SCAN}         & Jump      & 14670 & --- & 7706 \\
                              & Turn Left & 21890 & ---   & 1208 \\
                              & Length    & 11990 & ---   & 3920 \\
                              & MCD1      & 8365  & 1046  & 1045 \\
                              & MCD2      & 8365  & 1046  & 1045 \\
                              & MCD3      & 8365  & 1046  & 1045 \\
\midrule      
COGS                          & Gen       & 24K   & 12K & 12K  \\
\midrule      
\multirow{8}{*}{GeoQuery}     & Standard  & 600 & --- & 280 \\
                              & Template1 & 438 & 110 & 332 \\
                              & Template2 & 439 & 110 & 331 \\
                              & Template3 & 440 & 110 & 330 \\
                              & TMCD1     & 440 & 110 & 330 \\
                              & TMCD2     & 440 & 110 & 330 \\
                              & TMCD3     & 440 & 110 & 330 \\
                              & Length    & 440 & 110 & 330 \\
\midrule
\multirow{3}{*}{SMCalFlow-CS} & 8-shot    & 25412 & 1324 & 1325 \\
                              & 16-shot   & 25420 & 1324 & 1325 \\
                              & 32-shot   & 25436 & 1324 & 1325 \\
                        
\bottomrule
\end{tabular}}
\caption{Sizes of all datasets and splits.}
\label{tab:dataset_sizes}
\end{table}

\begin{table*}[t!]

\begin{center}
\scalebox{0.86}{
\begin{tabular}{ll}
\toprule
Dataset & Example \\
\midrule

\multirow{2}{*}{SCAN} & $x$: walk around right and jump thrice \\
                      & $y$: RTURN WALK RTURN WALK RTURN WALK RTURN WALK JUMP JUMP JUMP \\
\midrule
\multirow{2}{*}{COGS} & $x$: Camila gave a cake in a storage to Emma . \\
                      & $y$: give ( agent = Camila , theme = cake ( nmod . in = storage ) , recipient = Emma ) \\
\midrule
\multirow{2}{*}{GeoQuery} 
                      & $x$: what states border states that the m0 runs through\\
                      & $y$: answer ( intersection ( state , next$\_$to$\_$2 ( intersection ( state , traverse$\_$1 ( m0 ) ) ) ) )  \\
\midrule
\multirow{5}{*}{SMCalFlow-CS}
                      & $x$: create work meeting with my boss \\
                      & $y$: ( Yield :output ( CreateCommitEventWrapper :event (  CreatePreflightEventWrapper \\
                      & :constraint ( Constraint[Event] :attendees ( AttendeeListHasRecipient \\
                      & :recipient ( FindManager :recipient ( toRecipient ( CurrentUser ) ) ) ) \\
                      & :subject ( $\textrm{?= \#}$ ( String `` work meeting " ) ) ) ) ) ) \\
\bottomrule
\end{tabular}
}
\end{center}
\caption{Example inputs, $x$, and outputs, $y$.}
\label{tab:dataset_examples}
\end{table*}

\begin{table*}[ht]

\begin{center}
\scalebox{0.86}{
\begin{tabular}{ll}
\toprule
Dataset 
& Induced Rules \\
\midrule
\multirow{3}{*}{SCAN}

& $\mbox{NT} \rightarrow \langle \mbox{NT}_\boxnum{1}~\textrm{and}~\mbox{NT}_\boxnum{2} ,\mbox{NT}_\boxnum{1}~\mbox{NT}_\boxnum{2} \rangle$ \\
& $\mbox{NT} \rightarrow \langle \mbox{NT}_\boxnum{1}~\textrm{thrice} ,\mbox{NT}_\boxnum{1}~\mbox{NT}_\boxnum{1}~\mbox{NT}_\boxnum{1} \rangle$ \\
& $\mbox{NT} \rightarrow \langle \mbox{NT}_\boxnum{1}~\textrm{around right} ,\mbox{RTURN}~\mbox{NT}_\boxnum{1}~\mbox{RTURN}~\mbox{NT}_\boxnum{1}~\mbox{RTURN}~\mbox{NT}_\boxnum{1}~\mbox{RTURN}~\mbox{NT}_\boxnum{1} \rangle$ \\
\midrule
\multirow{3}{*}{COGS}
& $\mbox{NT} \rightarrow \langle \mbox{NT}_\boxnum{1}~\mbox{NT}_\boxnum{2}~\mbox{NT}_\boxnum{3}~\mbox{NT}_\boxnum{4} ,\mbox{NT}_\boxnum{2}~\textrm{( agent =}~\mbox{NT}_\boxnum{1}~\textrm{, theme =}~\mbox{NT}_\boxnum{3}~\textrm{, recipient =}~\mbox{NT}_\boxnum{4}~) \rangle$ \\
& $\mbox{NT} \rightarrow \langle \mbox{NT}_\boxnum{1}~\mbox{NT}_\boxnum{2}~\mbox{NT}_\boxnum{3},\mbox{NT}_\boxnum{1}~\textrm{( nmod .}~\mbox{NT}_\boxnum{2}~\textrm{ =}~\mbox{NT}_\boxnum{3}~\mbox{)}\rangle$ \\
& $\mbox{NT} \rightarrow \langle \mbox{NT}_\boxnum{1}, \mbox{a}~\mbox{NT}_\boxnum{1}\rangle$ \\
\midrule
\multirow{3}{*}{GeoQuery}
& $\mbox{NT} \rightarrow \langle \textrm{what}~\mbox{NT}_\boxnum{1}~\textrm{border}~\mbox{NT}_\boxnum{2}, \textrm{answer ( intersection (}~\mbox{NT}_\boxnum{1}~\textrm{ , next\_to\_2 ( }~\mbox{NT}_\boxnum{2}~\mbox{) ) )}\rangle$ \\
& $\mbox{NT} \rightarrow \langle \mbox{NT}_\boxnum{1}~\mbox{NT}_\boxnum{2}, \textrm{intersection (}~\mbox{NT}_\boxnum{1} ,~\mbox{NT}_\boxnum{2}~)\rangle$ \\
& $\mbox{NT} \rightarrow \langle \textrm{that}~\mbox{NT}_\boxnum{1}~\textrm{runs through}, \textrm{traverse\_1 ( }\mbox{NT}_\boxnum{1}~)\rangle$ \\
\midrule
\multirow{5}{*}{SMCalFlow-CS}
& $\mbox{NT} \rightarrow \langle \mbox{NT}_\boxnum{1}~\textrm{boss, FindManager :recipient ( }~\mbox{NT}_\boxnum{1}~) \rangle$ \\
& $\mbox{NT} \rightarrow \langle \mbox{NT}_\boxnum{1} \textrm{with}~\mbox{NT}_\boxnum{2},
\textrm{CreateCommitEventWrapper :event ( CreatePreflightEventWrapper}$ \\
& :constraint ( Constraint[Event] :attendees ( ( AttendeeListHasRecipient \\
& $\textrm{:recipient ( }~\mbox{NT}_\boxnum{2}~\textrm{ ) ) :subject ( ?~= \# ( }~\mbox{NT}_\boxnum{1}~\textrm{) ) ) )} \rangle$ \\
& $\mbox{NT} \rightarrow \langle \textrm{create}~\mbox{NT}_\boxnum{1},~\textrm{( Yield :output ( }~\mbox{NT}_\boxnum{1}~\textrm{ ) )} \rangle$ \\

\bottomrule
\end{tabular}
}
\end{center}
\caption{Examples of induced grammar rules for each example in Table~\ref{tab:dataset_examples}. Rules without non-terminals are omitted for brevity.}
\label{tab:grammar_examples}
\end{table*}

In this section we detail preprocessing for each dataset. Dataset sizes are reported in Table~\ref{tab:dataset_sizes}. We show examples of each dataset in Table~\ref{tab:dataset_examples}, with examples of the corresponding induced QCFG rules in Table~\ref{tab:grammar_examples}. For each dataset, we report exact match accuracy. We note that all datasets include English language data only; evaluating and extending our method for other languages is an important future direction. We use the same dataset preprocessing for the T5, T5+GECA, NQG-T5, and T5+CSL-Aug. results we report.

\paragraph{SCAN} We did not perform any preprocessing for SCAN. Grammar induction does not use any seed rules, and we do not assume a CFG defining valid output constructions, as the outputs consist of action sequences, not executable programs or logical forms.

\paragraph{COGS} For COGS, as QCFGs do not support logical variables~\cite{wong-mooney-2007-learning}, we mapped the original logical forms to a variable-free representation, with an example shown in Table~\ref{tab:dataset_examples}. 
The mapping is deterministic and reversible, and is akin to the use of other variable-free logical forms for semantic parsing such as FunQL~\cite{kate2005learning} or Lambda-DCS~\cite{liang2013lambda}.  An alternative but potentially more complex solution to handling logical variables in outputs would be to use an extension of SCFGs, such as $\lambda$-SCFG~\cite{wong-mooney-2007-learning}.

We define an output CFG based on the definition of this variable-free representation. To minimize the linguistic prior, we did not distinguish the types of primitives (e.g., nouns vs verbs); they all belong to the same CFG category. We use a set of seed rules of the form $NT \rightarrow \langle x',x \rangle$ where $x$ is a token found in a training output, and $x'$ is $x$ or an inflected form of $x$ found in a training input (e.g., for $x =$ ``sleep'', we add $NT \rightarrow \langle \text{sleep}, \text{sleep}\rangle$ and $NT \rightarrow \langle \text{slept}, \text{sleep}\rangle$). These $\langle x',x \rangle$ pairs were identified by running the IBM I alignment model~\cite{brown-etal-1993-mathematics} on the training data.

\paragraph{GeoQuery} 
We use the same variant of FunQL~\cite{kate2005learning} as \citet{shaw-etal-2021-compositional}, with entities replaced with placeholder values. We generate new length, template, and TMCD splits following the methodology of \citet{shaw-etal-2021-compositional}, so that we could evaluate our method on dev sets, which the original splits did not include. Specifically, for the length split, we randomly split the test set of the original length split into a dev set of 110 examples and a test set of 330 examples. To reduce variance, we created 3 new template and TMCD splits with different random seeds, with (approximately, in the case of template splits) 440 training examples, and 440 examples that are then randomly split into a 110 dev set and 330 test set. For the TMCD splits, we changed the atom constraint slightly, based on the error analysis in \citet{shaw-etal-2021-compositional} which found that a disproportionate amount of the errors on the TMCD test set were in cases where an ``atom'' was seen in only a single context during training. To create a fairer evaluation of compositional generalization, we strengthen the atom constraint such that every atom in the test set must be seen at least 2 times in the training set. Additionally, as several function symbols in FunQL can be used with and without arguments, and these usages are semantically quite different, we treat function symbols used with different numbers of arguments as different atoms. 

We define an output CFG based on the definition of the FunQL operators and the primitive types in the geobase database. We use a set of seed rules of the form $NT \rightarrow \langle x,x \rangle$ where $x$ occurs in both the input and output of a training example.

\paragraph{SMCalFlow-CS}
To construct SMCalFlow-CS, \citet{yin2021compositional} filtered out examples that require conversational context. We heuristically filtered out %
22 more training examples whose programs contain string literals that are not in the inputs.

We use the original LISP programs provided with the dataset as the output representation. We extract seed rules for string literals and numbers that are copied from inputs to outputs, such as person names and meeting subjects. We add 
5 seed rules with a single non-terminal on the input side that enable ``hallucinating'' various program fragments. We construct an output CFG based on the bracketing of LISP programs and a mapping of argument slots to nonterminals.

\section{Modeling Details}
\label{sec:appendix-model}

\subsection{QCFG Background and Notation}
\label{sec:appendix-model-qcfg}

Synchronous context-free grammars (SCFGs) have been used to model the hierarchical mapping between pairs of strings in areas such as compiler theory~\cite{aho1972theory} and multiple natural language tasks, e.g., machine translation~\cite{chiang2007hierarchical} and semantic parsing~\cite{wong-mooney-2006-learning,andreas-etal-2013-semantic}. 
SCFGs can be viewed as an extension of context-free grammars (CFGs) that \emph{synchronously} generate strings in what we will refer to as an input and output language. We write SCFG rules as $NT \rightarrow \langle \alpha,\beta \rangle $, where $NT$ is a non-terminal symbol, and $\alpha$ and $\beta$ are strings of non-terminal and terminal symbols. 

An SCFG rule can be viewed as two CFG rules, $NT \rightarrow \alpha$ and $NT \rightarrow \beta$, with a pairing between the occurrences of non-terminal symbols in $\alpha$ and $\beta$. This pairing is indicated by assigning each non-terminal in $\alpha$ and $\beta$ an index $\in \mathbb{N}$. Non-terminals sharing the same index are called \emph{linked}. Following convention, we denote the index for a non-terminal using a boxed subscript, e.g. $NT_{\boxnum{1}}$. 

\subsection{Model Parameterization Details}
\label{sec:appendix-model-parameterization}

Here we provide the complete definition of the $p_{\theta}(r|r_p,i)$ and $p_{\theta}(s|r_p,i)$ terms introduced in \S~\ref{sec:method-parameterization}:
\begin{align}
p_{\theta}(s|r_p,i) & = \frac{e^{\theta_{r_p,i,s}}}{\sum \nolimits_{s' \in \mathcal{S}} e^{\theta_{r_p,i,s'}} }\\
p_{\theta}(r|s) & =  \frac{e^{\theta_{s,r}}}{\sum \nolimits_{r' \in \mathcal{G}} e^{\theta_{s,r'}} } \label{eq:rgivens}
\end{align}
where the $\theta$s are scalar parameters. For SMCalFlow-CS, where the number of induced rules is large, we approximate the denominator in Eq.~\ref{eq:rgivens} by only considering rules used in derivations in the same batch during training.

\subsection{Grammar Induction Algorithm Details}
\label{sec:appendix-model-induction}

\begin{figure}[!t]
\begin{center}

\scalebox{0.9}{

\begin{tikzpicture} %
\begin{scope}[every node/.style={inner xsep=5pt, inner ysep=5pt}]

\tikzstyle{func} = [rectangle, rounded corners, text centered, draw=none, fill=none]

\node (n1) [func, align=center] {
$ \langle \mbox{NT}_\boxnum{1}, \mbox{NT}_\boxnum{1} \rangle$
};

\node (n2) [func, right=1.3cm of n1, align=center] {
$ \langle \alpha_a, \beta_a \rangle$
};

\node (n3) [func, below=0.8cm of n2, align=center] {
$ \langle \alpha_c, \beta_c \rangle$
};

\draw [-To] (n1) -- node[midway, below left] {$r_c$} (n3);
\draw [-To] (n1) -- (n2) node[midway,above] {$r_a$};
\draw [-To] (n2) -- (n3) node[midway,right] {$r_b$};

\end{scope}
\end{tikzpicture}

}
\\[0.3cm]
\scalebox{0.9}{

\begin{tikzpicture} %

\node (n1) [rectangle, draw=black, fill=none,  text width=7.0cm, inner xsep=5pt, inner ysep=5pt] {
{\centering{\textsc{Example}}\\[0.2cm]}

$r_a = \mbox{NT} \rightarrow \langle \mbox{NT}_\boxnum{1} $ and $\mbox{NT}_\boxnum{2} , \mbox{NT}_\boxnum{1}~ \mbox{NT}_\boxnum{2} \rangle$
\\
$r_b = \mbox{NT} \rightarrow \langle$jump$,$ JUMP$ \rangle$\\

$r_c = \mbox{NT} \rightarrow \langle$jump and $\mbox{NT}_\boxnum{1} ,$ JUMP $ \mbox{NT}_\boxnum{1} \rangle$

};

\end{tikzpicture}

}

\end{center}

\caption{
The arrows in the diagram denote expansion of a nonterminal with a rule. When the above ternary relation holds between $r_a$, $r_b$, and $r_c$, such as in the provided example, we will write $r_a \circ r_b \Rightarrow r_c$. The key sub-routine of our grammar induction algorithm, $\mathrm{UNIFY}(r_1, r_2)$, returns the set of rules $\{r_3 | r_2 \circ r_3 \Rightarrow r_1 \lor r_3 \circ r_2 \Rightarrow r_1\}$.
}
\label{fig:grammar_induction}
\end{figure}

In this section we describe the detailed implementation of the grammar induction algorithm introduced in \S~\ref{sec:induction}.

At each step, we process each rule $r_c$ in $\mathcal{G}$ in parallel. First, using a variant of the CKY algorithm, we check if we can just remove $r_c$ without violating the invariant that all examples in $\mathcal{D}$ can be derived by $\mathcal{G}$.
If so, we simply remove the rule $r_c$ as this will always decrease $L_\mathcal{D}(\mathcal{G})$.
Otherwise, we determine a set of candidate \emph{actions}, $A$, where an action $a \in A$ consists of a rule to add, $r_{add}$, and a set of rules to remove, $R_{remove}$. We determine $A$ using the $\mathrm{UNIFY}$ operation described in Figure~\ref{fig:grammar_induction}.
Specifically, we consider each rule returned by $\mathrm{UNIFY}(r_c, r')$ (where $r'$ is any other rule in $\mathcal{G}$) as a potential rule to add, $r_{add}$. The corresponding set $R_{remove}$ then consists of $r_c$ and any other rule that we determine can be removed if $r_{add}$ is added, without violating the above invariant.

If a CFG defining valid outputs is provided for the task, we ensure that the output string in $r_{add}$ can be generated by the given CFG, for some replacement of the nonterminal symbols with nonterminals from the output CFG, using a variant of the CKY algorithm.

Given these candidate actions, $A$, we select:
$$a_{max} = \underset{a \in A}{\argmax}~-L_\mathcal{D}(\mathrm{EXEC}(\mathcal{G}, a))$$
where $\mathrm{EXEC}(\mathcal{G}, a)$ is an operation that returns a new set of rules $(\mathcal{G} \cup r_{add}) \setminus R_{remove}$. 

We then aggregate over the actions, $a_{max}$, selected for each rule in $\mathcal{G}$, choosing an action only if it improves the objective. Each action is executed by setting $\mathcal{G} \gets \mathrm{EXEC}(\mathcal{G}, a_{max})$. The algorithm completes if no action was selected or if we reach a configurable number of steps.

We optionally partition the dataset into a configurable number of equally sized partitions based on example length. We then run the algorithm sequentially on each partition, starting with the partition containing the shortest examples. During initialization, we only add rules for examples in the first partition. We then add rules corresponding to examples in the next partition once the algorithm completes on the current partition.

\eat{
\subsection{Constrained Sampling}
\label{sec:appendix-model-sampling}

As described in \S~\ref{sec:augment}, if a CFG defining valid outputs is provided, we ensure that every $\langle x,y \rangle$ sampled from $\mathcal{G}$ contain a $y$ that is derivable given the provided CFG. While this can in theory be accomplished by sampling from $\mathcal{G}$ and then discarding examples where $y$ is not derivable by the provided CFG, this is undesirable for two reasons. First, this can be a computationally slow method to sample if many examples are discarded. Second, as longer examples may be more likely to be discarded under such a setting, this method can have undesirable effects on the resulting distribution of synthetic examples.

Instead, we use an alternative method. Given the constraint mentioned in the previous section, the output side of every rule in $\mathcal{G}$ can be derived given the target CFG, for some mapping between nonterminals. Therefore, we can determine a one-to-many mapping\linlu{I think it's actually many-to-many mapping? as 1 target NT can be mapped to >1 QCFG NT?} between the nonterminal tokens in $\mathcal{G}$ and the nonterminal symbols in the provided CFG. When sampling, we keep track of the set of possible nonterminals from the given CFG associated with the nonterminal from $\mathcal{G}$ being expanded, to ensure compatibility according to the given CFG.
\ps{This is probably confusing and can maybe be improved. Also not sure how important a point it is. At some point we probably need to refer the curious reader to the code.} \linlu{Perhaps just change the last sentence we sample from the intersection of G and target CFG? seem to be hard to explain without referring to the code. can maybe use a symbol to represent target CFG as it is mentioned in many different places}\ps{Pengcheng Yin specifically asked about this, but maybe its a pretty niche question that can be resolved once we release the code.} %
}

\subsection{Hyperparameters}

We performed a limited amount of hyperparameter tuning based on performance on development sets. As our goal is to develop models that generalize well across multiple types of distribution shifts, we strove to use the same hyperparameters for each split within a dataset. 

\paragraph{Grammar Induction} For grammar induction, we selected various configuration options by inspecting the data for each dataset, such as the maximum number of nonterminals in a rule and whether we allow repeated nonterminal indexes. We evaluated several configurations for $k_\alpha$ and $k_\beta$ in Eq.~\ref{eq:grammar_objective2} and the relative cost of terminal vs. nonterminal symbols referenced in Eq.~\ref{eq:grammar_objective}, which we will refer to here as $k_t$, during the development of our algorithm. The selected hyperparameters are listed in Table~\ref{tab:hyperparameters}.

\begin{table}[!t]
\begin{center}

\scalebox{0.85}{
\begin{tabular}{lcccc}
\toprule
Dataset & $k_\alpha$ & $k_\beta$ & $k_t$ & \# Partitions \\
\midrule
SCAN & 0 & 100 & 4 & 16 \\
COGS & 1 & 5 & 8 & 1 \\
GeoQuery & 4 & 16 & 8 & 1 \\
SMCalFlow-CS & 4 & 16 & 8 & 1 \\
\bottomrule
\end{tabular}
}
\caption{Hyperparameters for grammar induction.}
\label{tab:hyperparameters}
\end{center}
\end{table}

\paragraph{Parameteric Model} For the CSL parameteric model, we selected a learning rate from $[0.01, 0.05, 0.1]$. We selected a number of context states $|\mathcal{S}|$ from $[32, 64]$, %
except for SCAN where we analyzed a larger number of context states on the MCD splits, as discussed in Appendix~\ref{sec:appendix-analysis-states}. For SCAN we selected $|\mathcal{S}|=2$ for all splits, except for MCD1 and MCD3, where we found $|\mathcal{S}|=4$ to give more consistent performance on the dev sets. This was the only case where we used different hyperparameters for different splits of the same dataset.%

\paragraph{Sampling} We sampled $100,000$ synthetic examples for all datasets. We provide some analysis of the effect of this in Appendix~\ref{sec:appendix-analysis-unlabeled}. For COGS, we biased sampling to increase the number of longer examples (as discussed in \S~\ref{sec:augment}) by setting the bias $\delta=6$, and otherwise used $\delta=0$. We limited the maximum recursion depth to 5 for SCAN, 10 for SMCalFlow, and 20 for GeoQuery and COGS.

\paragraph{T5 Fine-Tuning} We started with the same configuration for fine-tuning T5 as \citet{shaw-etal-2021-compositional}. We similarly selected a learning rate from $[1e^{-3}, 1e^{-4}, 1e^{-5}]$ for each dataset. We use learning rate of $1e^{-3}$ for SCAN and SMCalFlow and $1e^{-4}$ for GeoQuery and COGS.

\subsection{Training Details}

We train the CSL model on 8 V100 GPUs, which takes less than 1.5 hours for all splits. We fine-tune T5 on 32 Cloud TPU v3 cores\footnote{\url{https://cloud.google.com/tpu/}} for 10,000 steps, which takes less than 6 hours for all splits.
\section{Additional Analysis}
\label{sec:appendix-analysis}

\begin{table*}[!t]
\begin{center}

\scalebox{0.8}{
\begin{tabular}{lccccccccccc}
\toprule
 & 
 & \multicolumn{3}{c}{{$x \in \mathcal{X}_{CSL}$}} 
 & \multicolumn{3}{c}{{$x \notin \mathcal{X}_{CSL}$}} 
 & \multicolumn{4}{c}{{All}} \\
 
 \cmidrule(lr){3-5} \cmidrule(lr){6-8} \cmidrule(lr){9-12}
 
Dataset & $\%^{\mathcal{X}_{CSL}}$ & T5 & CSL  & Aug. & T5 & CSL  & Aug. & T5 & CSL & Ens. & Aug. \\
\midrule
SCAN Jump & 100.0 & 99.5 & 100.0 & 99.7 & --- & --- & --- & 99.5 & 100.0 & 100.0 & 99.7  \\
SCAN Left & 100.0 & 62.0 & 100.0 & 100.0 & --- & --- & --- & 62.0 & 100.0 & 100.0 &  100.0  \\
SCAN Length & 100.0 & 14.4 & 100.0 & 99.2 & --- & --- & --- & 14.4 & 100.0 & 100.0 & 99.2  \\
SCAN MCD & 100.0 & 15.4 & 100.0 & 99.4 & --- & --- & --- & 15.4 & 100.0 & 100.0 & 99.4  \\

\midrule
COGS Gen. & 99.9 & 89.8 & 99.6 & 99.5 & 40.9 & 0.0 & 100.0 & 89.8 & 99.5 & 99.5 & 99.5 \\
\midrule
GeoQuery Std.   & 76.3 & 97.2 & 97.3 & 98.1 & 78.9 & 0.0 & 77.9 & 92.9 & 74.3 & 93.0 & 93.3 \\
GeoQuery Templ. & 61.0 & 93.1 & 96.6 & 97.1 & 71.6 & 0.0 & 76.9 & 84.8 & 58.9 & 86.9 & 89.3 \\
GeoQuery TMCD   & 44.3 & 88.4 & 90.3 & 93.9 & 53.8 & 0.0 & 59.9 & 69.2 & 39.9 & 70.0 & 74.9 \\
GeoQuery Length & 29.0 & 51.2 & 91.6 & 83.6 & 35.4 & 0.0 & 61.3 & 40.0 & 26.6 & 51.7 & 67.8 \\
\midrule
SMCalFlow-CS 8-S  & 30.5 & 96.0 & 85.6 & 95.5 & 79.8 & 0.0 & 78.3 & 84.7 & 26.1 & 81.6 & 83.5 \\
SMCalFlow-CS 8-C  & 6.6  & 52.3 & 79.6 & 72.7 & 33.4 & 0.0 & 50.1 & 34.7 & 5.3  & 36.5 & 51.6 \\
SMCalFlow-CS 16-S & 29.5 & 95.9 & 85.6 & 95.9 & 80.1 & 0.0 & 78.2 & 84.7 & 25.2 & 81.7 & 83.4 \\
SMCalFlow-CS 16-C & 11.6 & 59.7 & 84.4 & 84.4 & 42.7 & 0.0 & 58.4 & 44.7 & 9.8  & 47.5 & 61.4 \\
SMCalFlow-CS 32-S & 30.4 & 96.0 & 88.1 & 95.5 & 80.5 & 0.0 & 79.0 & 85.2 & 26.7 & 82.8 & 84.0 \\
SMCalFlow-CS 32-C & 13.0 & 74.4 & 87.2 & 88.4 & 56.7 & 0.0 & 67.8 & 59.0 & 11.3 & 60.6 & 70.4 \\
\bottomrule
\end{tabular}
}
\caption{Performance breakdown for all splits, including non-compositional and synthetic splits, in addition to those already presented in Table~\ref{tab:quadrant_analysis_small}.}
\label{tab:quadrant_analysis_full}
\end{center}
\end{table*}
\begin{table*}[th]
\begin{center}

\scalebox{0.8}{
\begin{tabular}{lccccccc}
\toprule
 & & \multicolumn{6}{c}{Context States, $|\mathcal{S}|$} \\ 
 \cmidrule(lr){3-8}

Split & Log prob. & 1 & 2 & 3 & 4 & 8 & 32 \\
\midrule
mcd1 (train) & $p(x,y)$ & -15.14 & -12.55 & -10.96 & -9.60 & -9.21 & -9.09 \\
mcd1 (train) & $p(y|x)$ & -0.72 & 0.00 & 0.00 & 0.00 & 0.00 & 0.00 \\
mcd1 (dev) & $p(x,y)$ & -16.49 & -15.04 & -17.47 & \bf -13.32 & -18.85 & -20.61 \\
mcd1 (dev) & $p(y|x)$ & -0.76 & 0.00 & 0.00 & 0.00 & 0.00 & 0.00 \\
\midrule
mcd2 (train) & $p(x,y)$ & -14.38 & -11.72 & -10.62 & -10.26 & -9.26 & -9.13 \\
mcd2 (train) & $p(y|x)$ & -0.58 & 0.00 & 0.00 & 0.00 & 0.00 & 0.00 \\
mcd2 (dev) & $p(x,y)$ & -17.60 & \bf -14.10 & -19.84 & -17.58 & -19.45 & -21.82 \\
mcd2 (dev) & $p(y|x)$ & -0.97 & 0.00 & -1.99 & -0.36 & 0.00 & -0.04 \\
\midrule
mcd3 (train) & $p(x,y)$ & -15.29 & -12.56 & -10.86 & -11.04 & -9.11 & -9.09 \\
mcd3 (train) & $p(y|x)$ & -0.75 & 0.00 & 0.00 & 0.00 & 0.00 & 0.00 \\
mcd3 (dev) & $p(x,y)$ & -16.33 & -14.96 & \bf -11.25 & -19.09 & -19.83 & -20.71 \\
mcd3 (dev) & $p(y|x)$ & -0.75 & 0.00 & 0.00 & -0.10 & 0.00 & 0.00 \\
\bottomrule
\end{tabular}
}
\caption{We compare the average log probability CSL assigns to examples from the SCAN MCD train and dev sets, for different numbers of context states, $|\mathcal{S}|$.} 
\label{tab:num_types}
\end{center}
\end{table*}

\subsection{Performance Breakdown}
\label{sec:appendix-analysis-breakdown}

We extend the performance breakdown analysis of \S\ref{sec:analysis} to all splits, with results reported in Table~\ref{tab:quadrant_analysis_full}. 

\subsection{Varying Context Sensitivity}
\label{sec:appendix-analysis-states}

Varying the number of context states $|\mathcal{S}|$ can vary the degree of context sensitivity in the CSL model. This can be important because we want our model to be able to accurately model $p(y|x)$, which we assume is shared between the source and target distributions, but we also want to sample new inputs $x$ that may have low probability under the source distribution due to the novel compositions they contain.

As a step towards understanding the trade-offs related to context sensitivity, we compute the average $\log p(x,y)$ and $\log p(y|x)$ according to CSL models with different number of context clusters $|\mathcal{S}|$. The results are reported in Table~\ref{tab:num_types}. The results also let us compare $\log p(x) = \log p(x,y) - \log p(y|x)$.

A constraint on the number of context states $|\mathcal{S}|$ is in some ways similar to a constraint on the number of nonterminal symbols in a conventional SCFG. Notably, for SCAN, writing a SCFG that unambiguously maps inputs to outputs requires 2 unique nonterminal symbols, and we observe that, similarly, $|\mathcal{S}| \geq 2$ is required to reach $100\%$ accuracy on the dev set. We also observe that while the models with larger $|\mathcal{S}|$ fit the training set better, the log likelihood of the dev sets is highest with $|\mathcal{S}|$ in the range between 2 and 4, indicating that the optimal place on the tradeoff curve is not at the extremes. We also note that there is some variance across the different splits for the optimal number of types, with some values leading to less than optimal modeling of $p(y|x)$.

It is also worth noting that, regardless of the number of context states, the structural conditional independence assumptions in our model can be too strong, harming the accuracy of modeling $p(y|x)$. For example, consider a rule for coordination, $\mbox{NT} \rightarrow \langle \mbox{NT}_\boxnum{1}~\textrm{and}~\mbox{NT}_\boxnum{2} ,\mbox{NT}_\boxnum{1}~\land~\mbox{NT}_\boxnum{2} \rangle$. In our model, we cannot condition the expansion of $\mbox{NT}_\boxnum{2}$ on the corresponding expansion of $\mbox{NT}_\boxnum{1}$ or the parent context in which the coordination rule is applied, in order to capture notions of type agreement.
Such limitations are similar to the limitations of conventional PCFGs to sufficiently model structural dependencies for syntactic parsing~\cite{klein-manning-2003-accurate}.

In general, better understanding and optimizing the trade-offs related to context sensitivity for compositional generalization is an important direction for future work.

\subsection{Comparing CSL and NQG}
\label{sec:appendix-analysis-nqg}

CSL and NQG of~\citet{shaw-etal-2021-compositional} vary across several dimensions, as the two systems use different grammar induction algorithms and different model parameterizations. Here we compare the two approaches across both dimensions independently. 

\paragraph{Grammar Induction}

As discussed in \S\ref{sec:induction}, the largest set of changes to the CSL algorithm from that of NQG were to improve the scalabilty of the induction algorithm, as both algorithms scale superlinearly in both dataset size and the length of input and output strings. The runtime of grammar induction on the GeoQuery standard split on a standard workstation CPU is around 15 minutes for NQG, and $<1$ minute for CSL. More importantly, we did not find it feasible to run NQG for SMCalFlow-CS, while CSL enables grammar induction to be completed within 10 hours with parallelization. 

CSL also supports QCFG rules with $>2$ nonterminals while NQG does not. We found allowing up to $4$ nonterminals can improve the coverage of induced grammars for COGS and SMCalFlow-CS, with some example rules shown in Table~\ref{tab:grammar_examples} in Appendix~\ref{sec:appendix-datasets}. Notably, for COGS, the induction algorithm of NQG induced a grammar that can only derive $64.9\%$ of the test set.

\paragraph{Model Parameterization}

\begin{table}[!t]
\begin{center}

\scalebox{0.75}{
\begin{tabular}{lcccc}
\toprule
 & \multicolumn{3}{c}{{\bf{\textsc{GeoQuery}}}} &  
 \multicolumn{1}{c}{{\bf\textsc{COGS}}} \\ 
 \cmidrule(lr){2-4} \cmidrule(lr){5-5}
 
Parsing Model & Templ. & TMCD & Len. & Gen. \\
\midrule
BERT + SpanLabel & 58.8 & 41.0 & 23.3 & 99.2 \\
CSL Gen. Model & 58.9 & 39.9 & 26.6 & 99.4 \\
\bottomrule
\end{tabular}
}
\caption{We compare the accuracy of the span-based BERT-Base model of NQG (BERT + SpanLabel) with that of the CSL generative model (when used as a parsing model) for the same induced grammar (induced using CSL), on compositional splits of GeoQuery and COGS.}
\label{tab:csl_vs_nqg}
\end{center}
\end{table}

We compare the performance of CSL's simple generative model with that of the span-based model of NQG which uses a BERT-Base encoder in Table~\ref{tab:csl_vs_nqg}. Both models use the \emph{same grammar} induced via the CSL algorithm.
Overall, the models perform comparably, despite CSL having far fewer parameters (e.g. for GeoQuery, CSL has only 51,200 parameters\footnote{For each rule and non-terminal token in the induced grammar, CSL has a number of parameters equal to the number of context states.} while NQG has over 110M parameters as it includes a BERT-Base encoder)%
, not leveraging pre-trained neural networks, and being a generative model to support sampling (in contrast to NQG which is a discriminative model). %
Incorporating pre-trained neural components into a model such as CSL could be a promising future direction. 

For SMCalFlow-CS, given a grammar induced by CSL, we found that it can be computationally infeasible to train a discriminative NQG model, due to the need to compute the partition function which sums over all possible derivations of the input. As a generative model, CSL avoids the need to compute a partition function during training.

\subsection{Comparison with GECA}
\label{sec:appendix-analysis-geca}

\paragraph{Hyperparameters} For SCAN, the reported results in Table \ref{tab:main_joint} use a window size of 1. Using a window size of 2 improves performance on the MCD split (24.9\% vs. 22.8\%) but hurts performance on the other splits. For GeoQuery, we used the default window size of 4 for the GeoQuery experiments. We attempted to run GECA on COGS and SMCalFlow-CS also using the default hyperparameters and did not find it to be computationally tractable. The algorithm's iteration over templates and fragments can become prohibitive for larger-scale datasets.

\paragraph{Analysis} From Table \ref{tab:main_joint} we see that augmenting the training data using CSL outperforms GECA across both synthetic and non-synthetic evaluations. GECA relies on the simple assumption that fragments are interchangeable if they appear in the same context. It is restricted by a pre-defined window size for fragments, and does not support recursion. 
Figure~\ref{fig:csl_vs_geca} compares differences in the sets of derivable synthetic examples for a notional set of training examples. In this case, CSL can derive a much larger set of recombinations by inducing the rule $\mbox{NT} \rightarrow \langle \mbox{NT}_\boxnum{1}$ and $\mbox{NT}_\boxnum{2},~\mbox{NT}_\boxnum{1}~ \mbox{NT}_\boxnum{2} \rangle$, which can be applied recursively.

\begin{figure}[!t]
\begin{center}

\scalebox{0.63}{

\begin{tikzpicture} %

\node (n1) [rectangle, draw=black, fill=none,  text width=11.0cm, inner xsep=5pt, inner ysep=5pt] {

{\centering{\textsc{Training Examples}}\\[0.3cm]}

$\langle$jump, JUMP$\rangle$ \\
$\langle$walk, WALK$\rangle$ \\ 
$\langle$jump and walk, JUMP WALK$\rangle$ \\[0.3cm]

{\centering{\textsc{Synthetic Examples (GECA)}}\\[0.3cm]}

$\langle$walk and walk, WALK WALK$\rangle$ \\ $\langle$jump and jump, JUMP JUMP$\rangle$ \\[0.3cm]

{\centering{\textsc{Synthetic Examples (CSL)}}\\[0.3cm]}

 $\langle$walk and walk, WALK WALK$\rangle$ \\
$\langle$jump and jump, JUMP JUMP$\rangle$ \\
$\langle$walk and jump, WALK JUMP$\rangle$ \\
$\langle$walk and walk and jump,
WALK WALK JUMP$\rangle$ \\
$\langle$walk and jump and walk and walk,
WALK JUMP WALK WALK$\rangle$ \\
$\cdots$ \\[0.3cm]

};

\end{tikzpicture}

}
\end{center}

\caption{We show the set of derivable synthetic examples for GECA (with window size = 2) and CSL, given an illustrative example of training examples. CSL can derive a significantly larger set of synthetic examples than GECA, and also assigns a probability to each derivable example.} 
\label{fig:csl_vs_geca}

\end{figure}

\subsection{Limitations of QCFGs}
\label{sec:qcfg-limitations}
The mapping from inputs to outputs in SCAN, COGS, and GeoQuery are all well supported by QCFGs. However, grammars were used to generate the data for SCAN and COGS, so this is perhaps not surprising. While GeoQuery inputs were written by humans, the distribution of queries in the dataset is influenced by the capabilities of the underlying execution engine based on logic programming; the dataset has a large number of nested noun phrases in inputs that map directly to nested FunQL clauses in outputs.

\paragraph{SMCalFlow} The induced grammars have relatively low coverage on SMCalFlow, as shown by $\%^{\mathcal{X}_{CSL}}$ in Table~\ref{tab:quadrant_analysis_full}, although they are still sufficient to improve the performance of T5. One reason for the low coverage is that inputs in SMCalFlow often reference specific names, locations, and meeting subjects, such as ``setup up a sales meeting with Sam and his manager'' where ``sales meeting'' and ``Sam'' must be copied to the output program as string literals. Sequence-to-sequence models with copy mechanisms or shared input-output vocabularies can handle such copying, but the QCFGs induced by our method do not support generalization to such novel tokens. Extending the method to support such string copying could significantly improve coverage.

Another reason for the low coverage is that the mismatch between the nesting of prepositional phrases in the input (e.g., ``at NT'' and ``with NT'') and the corresponding clauses in the output program tree makes it difficult to induce QCFG rules that enable recombination of different prepositional phrases in different contexts.

The induced QCFGs are also limited in other cases, such as their inability to ``distribute'' over groupings correctly. Since the training data only contains example such as $\langle$Jennifer and her boss, (person = ``Jennifer'') (FindManager (person= ``Jennifer''))$\rangle$, the induced rule NT $\rightarrow$ $\langle$NT and her boss, (person = ``NT'') (FindManager (person= ``NT''))$\rangle$, the induced grammar cannot correctly generate test examples like $\langle$Jennifer and Elli and their bosses, (person = ``Jennifer'') (person = ``Elli'') (FindManager (person= ``Jennifer'')) ( FindManager (person= ``Elli''))$\rangle$. 

\paragraph{CFQ} We also evaluated the feasibility of our approach to improve T5 performance on CFQ~\cite{keysers2019measuring}, a popular synthetic dataset for evaluating compositional generalization. We found it was challenging to induce QCFGs with reasonable coverage for CFQ. First, the SPARQL queries in CFQ contain variables, which are not well supported by QCFGs~\cite{wong-mooney-2007-learning}. Additionally, the mapping from queries to SPARQL in CFQ requires notions of commutativity (both ``M0 edited and directed M1'' and ``M0 directed and edited M1'' will be mapped to ``M0 ns:film.director.film M1 . M0 ns:film.editor.film M1'') and distributivity (edited in ``edited M1 and M2'' will appear twice in ``?x0 ns:film.editor.film M1 . ?x0 ns:film.editor.film M2'') that are also not well supported by QCFGs. Such limitations can potentially be partially overcome by desigining intermediate representations for CFQ ~\cite{furrer2020compositional, herzig2021unlocking}, but a complete solution likely requires an extension to the class of allowable rules in $\mathcal{G}$ beyond those a QCFG formalism supports, such as better support for variables~\cite{wong-mooney-2007-learning} and the ability to apply rewriting rules to generated output strings.

\subsection{Sampling Temperature}
\label{sec:appendix-analysis-temp}

\begin{table*}[!t]
\begin{center}

\scalebox{0.75}{
\begin{tabular}{lccccccccccccccc}
\toprule
 & \multicolumn{4}{c}{{\bf{\textsc{SCAN}}}}
 & \multicolumn{1}{c}{{\bf{\textsc{COGS}}}} 
 & \multicolumn{4}{c}{{\bf{\textsc{GeoQuery}}}} 
 & \multicolumn{6}{c}{{\bf\textsc{SMCalFlow-CS}}} \\
 \cmidrule(lr){2-5} \cmidrule(lr){6-6} \cmidrule(lr){7-10} \cmidrule(lr){11-16}
Temp. & Jump & Left & Len. & MCD & Gen. & Std. & Templ. & TMCD & Len. & 8-S & 8-C & 16-S & 16-C & 32-S & 32-C \\
\midrule
$\infty$ & 97.2 & 98.0 & 42.0 & 43.8 & 91.4 & 92.9 & 88.0 & 72.9 & 51.5 & 83.4 & 63.1 & 84.4 & 62.6 & 83.8 & 71.5 \\
10 & --- & --- & --- & --- & --- & --- & --- & --- & --- & 82.8 & 62.4 & 83.1 & 63.8 & 82.5 & 71.0 \\
1 & 
99.7 & 100.0 & 99.2 & 99.4 & 99.5 & 93.3 & 89.3 & 74.9 & 67.8 & 83.5 & 51.6 & 83.4 & 61.4 & 84.0 & 70.4 \\
\bottomrule
\end{tabular}
}
\caption{We compare sampling from CSL and uniform distribution (temperature = $\infty$) for the same induced grammar.} 
\label{tab:csl_vs_uni}
\end{center}
\end{table*}

In this section we study the impact of the parametric model on data augmentation. To do this, we consider varying the sampling temperature, applied to $\theta_{t,r}$ prior to normalization. We compare the accuracy of T5+CSL-Aug. for different temperatures for SMCalFlow-CS, and also with sampling from a uniform distribution rather than using the CSL parameteric model for all splits, which can be viewed as using temperature $= \infty$. 

Table~\ref{tab:csl_vs_uni} shows that using the parameteric model outperforms uniform sampling by a large margin on most splits. However, for SMCalFlow-CS, increasing the sampling temperature can lead to improved performance. To help understand why increasing temperature improves performance on the SMCalFlow-CS cross-domain splits, we computed the number of single-domain examples and cross-domain examples in the 100,000 sampled synthetic examples. Sampling from uniform distribution generates on average 17,764 cross-domain examples comparing with sampling from CSL which generates on average 1,114 cross-domain examples. The significant larger number of synthetic cross-domain examples might explain the improvement on cross-domain performance when increasing sampling temperature, especially on the 8-shot split, given the small number of cross-domain examples in the original training data.

\subsection{Semi-Supervised Learning}
\label{sec:appendix-analysis-unlabeled}

If we have unlabeled data from our target distribution, consisting of inputs only, we can incorporate this data when training our generative model in a straightforward way. Here we propose a new experiment to evaluate a method for semi-supervised learning that leverages such unlabeled examples. We assume that the unlabeled inputs from the development set are available during training.

\paragraph{Experiment Setting}
In this setting, conceptually, when optimizing $\theta$, we want to maximize both the likelihood of $p(x,y)$ for $\mathcal{D}$ and the marginal likelihood $p(x)$ for the unlabeled data. As the latter requires marginalizing over derivations for $x$ and all possible outputs, and this can be a large set to sum over, we approximate this by first labeling the unlabeled data following the inference procedure described in \S\ref{sec:inference-procedure} using the generative model trained only on $\mathcal{D}$, which can be interpreted as a {hard-EM} approach. During this process, we discard any unlabeled that cannot be derived given $\mathcal{G}$. We then re-train the generative model on both sets of data following the standard procedure, duplicating the ``unlabeled'' data a configurable number of times to achieve the desired ratio to the original labeled data.

We evaluate our CSL + T5-Aug. in this setting using \emph{unlabeled} data from the development set. We use the SCAN MCD splits as they have dev sets available. We also evaluate performance on the GeoQuery splits. We compare the performance with and without unlabeled data using 1,000 and 100,000 synthetic examples.

\begin{table}[!t]
\begin{center}

\scalebox{0.75}{
\begin{tabular}{ccccccc}
\toprule
 & & \multicolumn{4}{c}{{\bf{\textsc{GeoQuery}}}} &  
 \multicolumn{1}{c}{{\bf\textsc{SCAN}}} \\ 
 \cmidrule(lr){3-6} \cmidrule(lr){7-7}
 
Semi. & Example \# & Std. & Templ. & TMCD & Len. & MCD \\
\midrule
           & 1K & 92.9 & 89.3 & 72.5 & 66.4 & 63.8 \\
\checkmark & 1K & 93.2 & 89.3 & 74.4 & 67.9 & 88.9 \\
           & 100K & 93.3 & 89.3 & 74.9 & 67.8 & 99.4 \\
\checkmark & 100K & 93.2 & 89.4 & 75.6 & 67.3 & 99.4 \\
\bottomrule
\end{tabular}
}
\caption{We compare the accuracy of T5 + CSL using data augmentation with and without unlabeled data. } 
\label{tab:semi_supervised}
\end{center}
\end{table}

\paragraph{Results}
Results are reported in Table~\ref{tab:semi_supervised}. Incorporating unlabeled examples leads to improvements when sampling only 1,000 examples, but leads to minimal improvements when sampling 100,000 examples. We did not find positive results based on initial experiments for SMCalFlow-CS, likely due to the low coverage of the induced grammars on the target examples (see Table~\ref{tab:quadrant_analysis_full}), as the method we evaluated cannot leverage unlabeled examples that are not covered by the induced grammar. 

\subsection{GeoQuery Variance}
\label{sec:appendix-analysis-geoquery}
\begin{table}[!t]
\begin{center}

\scalebox{0.8}{
\begin{tabular}{lcc}
\toprule
Split & Mean & Stdev. \\ 
\midrule
Std. & 93.3 & 0.2 \\
Templ.1 & 92.5 & 0.3 \\
Templ.2 & 88.1 & 0.8 \\
Templ.3 & 87.2 & 0.6 \\
TMCD1 & 77.2 & 1.1 \\
TMCD2 & 71.3 & 0.6 \\
TMCD3 & 76.3 & 0.2 \\
Len. & 67.8 & 0.3 \\
\bottomrule
\end{tabular}
}
\caption{The mean and standard deviation of 3 runs for T5+CSL-Aug. on GeoQuery dataset.}
\label{tab:geoquery_runs}
\end{center}
\end{table}

The variance of T5+CSL-Aug. for GeoQuery is reported in Table~\ref{tab:geoquery_runs}.

\subsection{SMCalFlow-CS Error Analysis}
\label{sec:appendix-analysis-smcalflow}
\begin{table*}[!ht]
\begin{center}

\scalebox{0.85}{

\begin{tabular}{p{1.15\linewidth}}
\toprule

\emph{\bf Train Source:} Can you create an Meeting for Saturday 1 : 00 pm \\
\emph{\bf Train Target:} ( Yield :output ( CreateCommitEventWrapper :event ( CreatePreflightEventWrapper :constraint ( Constraint[Event] :start ( ?= ( DateAtTimeWithDefaults :date ( NextDOW :dow \# ( DayOfWeek " SATURDAY " ) ) :time ( NumberPM :number \# ( Number 1 ) ) ) ) \textcolor{red}{:subject ( ?= \# ( String " Meeting " ) )} ) ) ) ) \\
\\
\emph{\bf Dev Source:} Schedule a meeting on Thursday at 8 : 30 AM . \\
\emph{\bf Dev Target:} ( Yield :output ( CreateCommitEventWrapper :event ( CreatePreflightEventWrapper :constraint ( Constraint[Event] :start ( ?= ( DateAtTimeWithDefaults :date ( NextDOW :dow \# ( DayOfWeek " THURSDAY " ) ) :time ( HourMinuteAm :hours \# ( Number 8 ) :minutes \# ( Number 30.0 ) ) ) ) ) ) ) ) \\
\emph{\bf Dev Prediction:} ( Yield :output ( CreateCommitEventWrapper :event ( CreatePreflightEventWrapper :constraint ( Constraint[Event] :start ( ?= ( DateAtTimeWithDefaults :date ( NextDOW :dow \# ( DayOfWeek " THURSDAY " ) ) :time ( HourMinuteAm :hours \# ( Number 8 ) :minutes \# ( Number 30.0 ) ) ) ) \textcolor{red}{:subject ( ?= \# ( String " meeting " ) )} ) ) ) ) \\
\midrule
\emph{\bf Train Source:} Schedule 3 pm tentative shareholders huddle \\
\emph{\bf Train Target:} ( Yield :output ( CreateCommitEventWrapper :event ( CreatePreflightEventWrapper :constraint ( Constraint[Event] :start ( ?= ( NextTime :time ( NumberPM :number \# ( Number 3 ) ) ) ) \textcolor{red}{:subject ( ?= \# ( String " tentative shareholders huddle " ) )} ) ) ) ) \\
\\
\emph{\bf Dev Source:} Schedule 3 pm tentative shareholders huddle \\
\emph{\bf Dev Target:} ( Yield :output ( CreateCommitEventWrapper :event ( CreatePreflightEventWrapper :constraint ( Constraint[Event] \textcolor{red}{:showAs ( ?= \# ( ShowAsStatus " Tentative " ) )} :start ( ?= ( NextTime :time ( NumberPM :number \# ( Number 3 ) ) ) ) \textcolor{red}{:subject ( ?= \# ( String " shareholders huddle " ) )} ) ) ) ) \\
\emph{\bf Dev Prediction:} ( Yield :output ( CreateCommitEventWrapper :event ( CreatePreflightEventWrapper :constraint ( Constraint[Event] :start ( ?= ( NextTime :time ( NumberPM :number \# ( Number 3 ) ) ) ) \textcolor{red}{:subject ( ?= \# ( String " tentative shareholders huddle " ) )} ) ) ) )
 \\
\midrule
\emph{\bf Dev Source:} create football game on tuesday at 8 \\
\emph{\bf Dev Target:} ( Yield :output ( CreateCommitEventWrapper :event ( CreatePreflightEventWrapper :constraint ( Constraint[Event] :start ( ?= ( DateAtTimeWithDefaults :date ( NextDOW :dow \# ( DayOfWeek " TUESDAY " ) ) :time ( \textcolor{red}{NumberPM} :number \# ( Number 8 ) ) ) ) :subject ( ?= \# ( String " football game " ) ) ) ) ) ) \\
\emph{\bf Dev Prediction:} ( Yield :output ( CreateCommitEventWrapper :event ( CreatePreflightEventWrapper :constraint ( Constraint[Event] :start ( ?= ( DateAtTimeWithDefaults :date ( NextDOW :dow \# ( DayOfWeek " TUESDAY " ) ) :time ( \textcolor{red}{NumberAM} :number \# ( Number 8 ) ) ) ) :subject ( ?= \# ( String " football game " ) ) ) ) ) )
 \\
\bottomrule
\end{tabular}
}

\caption{Example prediction errors for T5+CSL-Aug. and their closest training example if any for the SMCalFlow-CS dataset.}
\label{tab:smcalflow_errors}

\end{center}
\end{table*}

We sampled 20 prediction errors for T5+CSL-Aug. from single-domain and cross-domain development sets respectively. We found a large number of errors are due to ambiguous and inconsistent annotations. Table~\ref{tab:smcalflow_errors} shows some examples of such errors. First, the subject string is determined inconsistently for training and testing examples. Second, the same source can be mapped to different targets which express the same meaning. Third, some examples require additional context to generate the correct output. Among the errors we sample, around 60\% of single-domain errors and around 35\% of cross-domain errors fall into these three types. In addition, for the cross-domain examples, T5+CSL-Aug. sometimes struggles with nesting programs in a correct way when examples require querying an org chart for more than one people as discussed in Appendix~\ref{sec:qcfg-limitations}.

\end{document}